\documentclass{article}

\usepackage{microtype}
\usepackage{graphicx}
\usepackage{booktabs} 

\usepackage{hyperref}

\usepackage{amsfonts}       
\usepackage{url}
\usepackage{amsmath}
\usepackage{amsthm}
\usepackage{wrapfig}
\usepackage{mathtools}
\usepackage{tikz}
\usepackage{subfig}
\usepackage{algorithmic}
\usepackage{subfig}
\usepackage{footmisc}

\newcounter{thm_counter}
\newcounter{lem_counter}
\newcounter{pro_counter}

\newcounter{ass_counter}

\newcounter{rmk_counter}
\newcounter{con_counter}
\newtheorem{theorem}[thm_counter]{Theorem}
\newtheorem{proposition}[pro_counter]{Proposition}
\newtheorem{lemma}[lem_counter]{Lemma}

\newtheorem{assumption}[ass_counter]{Assumption}

\newtheorem{remark}[rmk_counter]{Remark}
\newtheorem{condition}[con_counter]{Condition}

\newcommand{\mop}{\hat{\mathcal{T}}}
\newcommand{\qop}{\tilde{\mathcal{T}}}
\newcommand{\ns}{{|\mathcal{S}|}}
\newcommand{\dy}{{d_\mathcal{Y}}}
\newcommand{\py}{P_\mathcal{Y}}

\newcommand{\p}[1]{{#1}^\prime}

\usepackage{mathtools}
\usepackage{mathrsfs}
\mathtoolsset{showonlyrefs}
\newcommand{\E}{\mathbb{E}}
\newcommand{\R}{\mathbb{R}}

\usepackage[inline,ignoremode]{trackchanges}
\addeditor{lb}

\addeditor{sz}

\addeditor{yh}

\usepackage[accepted]{icml2020}

\icmltitlerunning{Convergent Two-timescale Off-Policy Actor-Critic}

\begin{document}
\twocolumn[
\icmltitle{Provably Convergent Two-Timescale Off-Policy Actor-Critic \\ with Function Approximation}



\icmlsetsymbol{equal}{*}

\begin{icmlauthorlist}
\icmlauthor{Shangtong Zhang}{ox}
\icmlauthor{Bo Liu}{au}
\icmlauthor{Hengshuai Yao}{hw}
\icmlauthor{Shimon Whiteson}{ox}
\end{icmlauthorlist}

\icmlaffiliation{ox}{University of Oxford}
\icmlaffiliation{au}{Auburn University}
\icmlaffiliation{hw}{Huawei Technologies}

\icmlcorrespondingauthor{Shangtong Zhang}{\mbox{shangtong.zhang@cs.ox.ac.uk}}

\icmlkeywords{Reinforcement Learning, Gradient Temporal Difference(TD) Learning, Actor-Critic, Emphatic TD, Off-poliyc Control}

\vskip 0.3in
]



\printAffiliationsAndNotice{}  

\begin{abstract}
We present the first provably convergent two-timescale off-policy actor-critic algorithm (COF-PAC) with function approximation.
Key to COF-PAC is the introduction of a new critic, the \emph{emphasis critic}, 
which is trained via Gradient Emphasis Learning (GEM), 
a novel combination of the key ideas of Gradient Temporal Difference Learning and Emphatic Temporal Difference Learning.
With the help of the emphasis critic and the canonical value function critic, 
we show convergence for COF-PAC,
where the critics are linear, and the actor can be nonlinear. 
\end{abstract}
\setcounter{footnote}{3}
\section{Introduction}
The policy gradient theorem and the corresponding actor-critic algorithm \citep{sutton2000policy,konda2002thesis} have recently enjoyed great success in various domains,
e.g., defeating the top human player in Go \citep{silver2016mastering}, achieving human-level control in Atari games \citep{mnih2016asynchronous}.
However, the canonical actor-critic algorithm is on-policy and hence suffers from significant data inefficiency (e.g., see \citet{mnih2016asynchronous}).
To address this issue, \citet{degris2012off} propose the Off-Policy Actor-Critic (Off-PAC) algorithm.
Off-PAC has been extended in various ways, e.g., off-policy Deterministic Policy Gradient (DPG, \citealt{silver2014deterministic}), 
Deep Deterministic Policy Gradient (DDPG, \citealt{lillicrap2015continuous}),
Actor Critic with Experience Replay (ACER, \citealt{wang2016sample}),
off-policy Expected Policy Gradient (EPG, \citealt{ciosek2017expected}),
TD3 \citep{fujimoto2018addressing},
 and
IMPALA \citep{espeholt2018impala}.
Off-PAC and its extensions have enjoyed great empirical success as the canonical on-policy actor-critic algorithm.
There is, however, a theoretical gap between the canonical on-policy actor-critic and Off-PAC.
Namely, on-policy actor-critic has a two-timescale convergent analysis under function approximation \citep{konda2002thesis},
but Off-PAC is convergent only in the tabular setting \citep{degris2012off}.
While there have been several attempts to close this gap \citep{imani2018off,maei2018convergent,zhang2019generalized,liu2019off},
none of them is convergent under function approximation without imposing strong assumptions (e.g., assuming the critic converges).

In this paper, we close this long-standing theoretical gap via the Convergent Off-Policy Actor-Critic (COF-PAC) algorithm,
the first provably convergent two-timescale off-policy actor-critic algorithm.
COF-PAC builds on Actor-Critic with Emphatic weightings (ACE, \citealt{imani2018off}), which reweights Off-PAC updates with \emph{emphasis} through the \emph{followon trace} \citep{sutton2016emphatic}.
The emphasis accounts for off-policy learning by adjusting the state distribution, and the followon trace approximates the emphasis (see \citealt{sutton2016emphatic}).\footnote{We use emphasis to denote the limit of the expectation of the followon trace, which is slightly different from \citet{sutton2016emphatic} and is clearly defined in the next section.} 
Intuitively, 
estimating the emphasis of a state using the followon trace is similar to estimating the value of a state using a single Monte Carlo return.
Thus it is not surprising that the followon trace can have unbounded variance \citep{sutton2016emphatic} and large emphasis approximation error,
 complicating the convergence analysis of ACE.

Instead of using the followon trace, 
we propose a novel stochastic approximation algorithm,
Gradient Emphasis Learning (GEM), to approximate the emphasis in COF-PAC,
inspired by Gradient TD methods (GTD, \citealt{sutton2009convergent,sutton2009fast}),
Emphatic TD methods (ETD, \citealt{sutton2016emphatic}), and reversed TD methods \citep{wang2007dual,wang2008stable,hallak2017consistent,gelada2019off}.
We prove the almost sure convergence of GEM, as well as other GTD-style algorithms, with linear function approximation under a slowly changing target policy.
With the help of GEM, 
we prove the convergence of COF-PAC, 
where the policy parameterization can be nonlinear, and the convergence level is the same as the on-policy actor-critic \citep{konda2002thesis}.



\section{Background}
We use $||x||_\Xi \doteq \sqrt{x^\top \Xi x}$ to denote the norm induced by a positive definite matrix $\Xi$, which induces the matrix norm $||A||_\Xi \doteq \sup_{||x||_\Xi = 1} || Ax ||_\Xi$.
To simplify notation, we write $||\cdot||$ for $||\cdot||_I$, where $I$ is the identity matrix.
All vectors are column vectors. 
We use ``0'' to denote an all-zero vector and an all-zero matrix when the dimension can be easily deduced from the context, and similarly for ``1''.
When it does not confuse, we use vectors and functions interchangeably.
Proofs are in the appendix.

We consider an infinite-horizon Markov Decision Process (MDP) with a finite state space $\mathcal{S}$ with $|\mathcal{S}|$ states,
a finite action space $\mathcal{A}$ with $|\mathcal{A}|$ actions, a transition kernel $p: \mathcal{S} \times \mathcal{S} \times \mathcal{A} \rightarrow [0, 1]$, a reward function $r: \mathcal{S} \times \mathcal{A} \times \mathcal{S} \rightarrow \mathbb{R}$, and a discount factor $\gamma \in [0, 1)$.
At time step $t$, an agent at a state $S_t$ takes an action $A_t$ according to $\mu(\cdot | S_t)$, 
where
$\mu: \mathcal{A} \times \mathcal{S} \rightarrow [0, 1]$ is a \emph{fixed behavior policy}.
The agent then proceeds to a new state $S_{t+1}$ according to $p(\cdot | S_t, A_t)$ and
gets a reward $R_{t+1} \doteq r(S_t, A_t, S_{t+1})$.
In the off-policy setting, the agent is interested in a \emph{target policy} $\pi$.
We use $G_t \doteq \sum_{k=1}^\infty \gamma^{k-1}R_{t+k}$
to denote the return at time step $t$ when following $\pi$ instead of $\mu$.
Consequently, we define the state value function $v_\pi$ and the state action value function $q_\pi$ as
$v_\pi(s) \doteq \mathbb{E}_\pi[G_t | S_t = s]$ and 
$q_\pi(s, a) \doteq \mathbb{E}_\pi[G_t | S_t = s, A_t = a]$.
We use $\rho(s, a) \doteq \frac{\pi(a | s)}{\mu(a | s)}$ to denote the importance sampling ratio and define $\rho_t \doteq \rho(S_t, A_t)$ (Assumption~\ref{assum:mdp} below ensures $\rho$ is well-defined).
We sometimes write $\rho$ as $\rho_\pi$ to emphasize its dependence on $\pi$.

\textbf{Policy Evaluation: }
We consider linear function approximation for policy evaluation.
Let $x: \mathcal{S} \rightarrow \mathbb{R}^{K_1}$ be the state feature function, 
and $\tilde{x}: \mathcal{S} \times \mathcal{A} \rightarrow \mathbb{R}^{K_2}$ be the state-action feature function.
We use $X \in \mathbb{R}^{\ns \times K_1}$ and $\tilde{X} \in \mathbb{R}^{N_{sa} \times K_2} (N_{sa} \doteq |\mathcal{S}| \times |\mathcal{A}|)$ to denote feature matrices,
where each row of $X$ is $x(s)$ and each row of $\tilde{X}$ is $\tilde{x}(s, a)$.
We use as shorthand that $x_t \doteq x(S_t), \tilde{x}_t \doteq \tilde{x}(S_t, A_t)$.
Let $d_\mu \in \mathbb{R}^{\ns} $ be the stationary distribution of $\mu$; 
we define $\tilde{d}_\mu \in \mathbb{R}^{N_{sa}}$ where $\tilde{d}_\mu(s, a) \doteq d_\mu(s) \mu(a|s)$. 
We define $D \doteq diag(d_\mu) \in \mathbb{R}^{\ns \times \ns}$ and $\tilde{D} \doteq diag(\tilde{d}_\mu) \in \mathbb{R}^{N_{sa} \times N_{sa}}$.
Assumption~\ref{assum:mdp} below ensures $d_\mu$ exists and $D$ is invertible, as well as $\tilde{D}$.
Let $P_\pi \in \mathbb{R}^{\ns \times \ns}$ be the state transition matrix and $\tilde{P}_\pi \in \mathbb{R}^{N_{sa} \times N_{sa}}$ be the state-action transition matrix, i.e., $P_\pi(s, s^\prime) \doteq \sum_a \pi(a|s)p(s^\prime|s, a), \tilde{P}_\pi((s, a), (s^\prime, a^\prime)) \doteq p(s^\prime|s, a)\pi(a^\prime | s^\prime)$.
We use $v\doteq X\nu, q\doteq \tilde{X}u$ to denote estimates for $v_\pi, q_\pi$ respectively, 
where $\nu, u$ are learnable parameters.

We first consider GTD methods.
For a vector $v \in \mathbb{R}^{\ns}$, 
we define a projection $\Pi v \doteq Xy^*, y^* \doteq \arg\min_y ||Xy - v||_D^2$.
We have $\Pi = X(X^\top D X)^{-1}X^\top D$ (Assumption~\ref{assu:non_singular} below ensures the existence of $(X^\top D X)^{-1}$). 
Similarly, for a vector $q \in \mathbb{R}^{N_{sa}}$, we define a projection $\tilde{\Pi} \doteq \tilde{X}(\tilde{X}^\top \tilde{D} \tilde{X})^{-1}\tilde{X}^\top \tilde{D}$.
The value function $v_\pi$ is the unique fixed point of the Bellman operator $\mathcal{T}: \mathcal{T}v \doteq r_\pi + \gamma P_\pi v$ where $r_\pi(s) \doteq \sum_{a, \p{s}} \pi(a|s) p(\p{s}|s, a) r(s, a, \p{s})$.
Similarly, $q_\pi$ is the unique fixed point for the operator $\qop: (\qop q)(s, a) \doteq \tilde{r} + \gamma \tilde{P}_\pi q$, 
where $\tilde{r} \in \R^{N_{sa}}$ and $\tilde{r}(s, a) \doteq \sum_{\p{s}} p(\p{s}|s, a) r(s, a, \p{s})$.
GTD2 \citep{sutton2009fast} learns the estimate $v$ for $v_\pi$, 
by minimizing $||\Pi\mathcal{T}v - v||_D^2$.
GQ(0) \citep{maei2011gradient} learns the estimate $q$ for $q_\pi$
by minimizing $||\tilde{\Pi} \tilde{\mathcal{T}} q - q||_{\tilde{D}}^2$.
Besides GTD methods, ETD methods are also used for off-policy policy evaluation.
ETD(0) updates $\nu$ as
\begin{align}
\label{eq:update_M_t}
M_t &\doteq i(S_t) + \gamma \rho_{t-1}M_{t-1}, \\
\label{eq:etd_update}
\nu_{t+1} &\doteq \nu_t + \alpha M_t \rho_t (R_{t+1} + \gamma x_{t+1}^\top \nu_t - x_t^\top \nu_t)x_t^\top,
\end{align}
where $\alpha$ is a step size, $M_t$ is the followon trace, and $i: \mathcal{S} \rightarrow [0, \infty)$ is the interest function reflecting the user's preference for different states \citep{sutton2016emphatic}. 
The interest is usually set to 1 for all states \citep{sutton2016emphatic},
meaning all states are equally important.
But if we are interested in learning an optimal policy for only a subset of states (e.g., the initial states),
we can set the interest to 1 for those states and to 0 otherwise \citep{white2017unifying}.
The interest function can also be regarded as a generalization of the initiation set in the option framework \citep{white2017unifying}. 
We refer the reader to \citet{white2017unifying} for more usage of the interest function.

\textbf{Control: }
Off-policy actor-critic methods \citep{degris2012off,imani2018off} aim to 
maximize the excursion objective
$J(\pi) \doteq \sum_s d_\mu(s) i(s) v_\pi(s)$
by adapting the target policy $\pi$.
We assume $\pi$ is parameterized by $\theta \in \mathbb{R}^K$,
and use $\theta, \pi, \pi_\theta$ interchangeably in the rest of this paper when it does not confuse.
All gradients are taken w.r.t.\ $\theta$ unless otherwise specified.
According to the off-policy policy gradient theorem \citep{imani2018off}, the policy gradient is
$\nabla J(\theta) = \sum_s \bar{m}(s) \sum_a q_\pi(s, a) \nabla \pi (a|s)$,
where $\bar{m} \doteq (I - \gamma P_\pi^\top)^{-1}D i \in \mathbb{R}^{\ns}$.
We rewrite $\bar{m}$ as $D D^{-1} (I - \gamma P_\pi^\top)^{-1}D i$ and define 
\begin{align}
\textstyle
    m_\pi \doteq D^{-1} (I - \gamma P_\pi^\top)^{-1}D i. 
\end{align}
We therefore have $\bar{m} = D m_\pi$, yielding 
\begin{align}
\textstyle
    \label{eq:grad_J_theta}
    \nabla J(\theta) = \E_{s \sim d_\mu, a \sim \mu(\cdot |s)} [m_\pi(s) \psi_\theta(s, a) q_\pi(s, a)], \quad
\end{align}
where 
$\psi_\theta(s, a) \doteq \rho_\theta(s, a)\nabla \log \pi(a|s) \in \mathbb{R}^K$.
We refer to $m_\pi$ as the \emph{emphasis} in the rest of this paper.
To compute $\nabla J(\theta)$, we need $m_\pi$ and $q_\pi$, to which we typically do not have access.
\citet{degris2012off} ignore the emphasis $m_\pi$ and update $\theta$ as $\theta_{t+1} \leftarrow \theta_t + \alpha \rho_t q_\pi(S_t, A_t) \nabla \log \pi(A_t|S_t)$ in Off-PAC,
which is theoretically justified only in the tabular setting.\footnote{See Errata in \citet{degris2012off}}
\citet{imani2018off} approximate $m_\pi(S_t)$ with the followon trace $M_t$, yielding the ACE update
$\theta_{t+1} \leftarrow \theta_t + \alpha M_t \rho_t q_\pi(S_t, A_t) \nabla \log \pi(A_t | S_t)$.
Assuming $\lim_{t \rightarrow \infty}\mathbb{E}_\mu[M_t | S_t = s]$
exists and $\pi$ is fixed, \citet{sutton2016emphatic} show that 
$\lim_{t \rightarrow \infty}\mathbb{E}_\mu[M_t | S_t = s] = m_\pi(s)$.
The existence of this limit is later established in Lemma 1 in \citet{zhang2019generalized}. 

\subsection{Assumptions}

\begin{assumption}
    \label{assum:mdp}
    The Markov chain induced by the behavior policy $\mu$ is ergodic,
    and $\forall (s, a), \, \mu(a | s) > 0$.
\end{assumption}
\begin{assumption}
    \label{assu:non_singular}
    The matrices 
    $C \doteq X^\top D X, \tilde{C} \doteq \tilde{X}^\top \tilde{D} \tilde{X} $ are nonsingular.
\end{assumption}

\begin{assumption}
    \label{assu:policy_params}
    There exists a constant $C_0 < \infty$ such that $\forall (s, a, \theta, \bar{\theta}),$
    \begin{align}
        &||\psi_\theta(s, a)|| \leq C_0, || \nabla \psi_\theta(s, a)|| \leq C_0 \\
        &|\pi_{\theta}(a|s) - \pi_{\bar{\theta}}(a|s)|\leq C_0 ||\theta - \bar{\theta}||, \\
        &||\psi_{\theta}(s, a) - \psi_{\bar{\theta}}(s, a)|| \leq C_0 ||\theta - \bar{\theta}||.
    \end{align}
\end{assumption}
\begin{remark}
The nonsingularity in Assumption~\ref{assu:non_singular} is commonly assumed in GTD methods \citep{sutton2009convergent, sutton2009fast, maei2011gradient} and can be satisfied by using linearly independent features.
Assumption~\ref{assu:policy_params} contains common assumptions for policy parameterization akin to those of \citet{sutton2000policy,konda2002thesis}.
\end{remark}
\begin{lemma}
    \label{lem:boundess_policy_params}
    Under Assumptions (\ref{assum:mdp}, \ref{assu:policy_params}), there exists a constant $C_1 < \infty$ such that $\forall (\theta, \bar{\theta})$
    \begin{align}
    &||\nabla J(\theta)|| \leq C_1, \,
    ||\nabla J(\theta) - \nabla J(\bar{\theta})|| \leq C_1||\theta - \bar{\theta}||, \\
    &||\textstyle{\frac{\partial^2 J(\theta)}{\partial \theta_i \partial \theta_j}}|| \leq C_1.
    \end{align}
\end{lemma}
\begin{lemma}
\label{lem:p_pi_norm}
Under Assumption~\ref{assum:mdp},
$||P_\pi||_D = ||D^{-1}P_\pi^\top D||_D$
\end{lemma}

\section{Gradient Emphasis Learning}

\textbf{Motivation: }
The followon trace $M_t$ has two main disadvantages.
First, when we use $M_t$ to approximate $m_\pi(S_t)$ (e.g., in ACE), 
the approximation error tends to be large.
$M_t$ is a random variable and
although its conditional expectation $\E_\mu[M_t|S_t=s]$ converges to $m_\pi(S_t)$ under a fixed target policy $\pi$,
$M_t$ itself can have unbounded variance \citep{sutton2016emphatic},
indicating the approximation error $|M_t - m_\pi(S_t)|$ can be unbounded.
Moreover, in our actor-critic setting, where $\pi$ keeps changing, 
it is not clear whether this convergence holds or not.
Theoretically, this large approximation error may preclude a convergent analysis for ACE.
Empirically, this large variance makes ETD hard to use in practice.
For example, as pointed out in~\citet{sutton2018reinforcement}, ``it is nigh impossible to get consistent results in computational experiments'' (for ETD) in Baird's counterexample \citep{baird1995residual}, a common off-policy learning benchmark.

Second, it is hard to query the emphasis $m_\pi(s)$ for a given state $s$ using the followon trace.
As $M_t$ is only a scalar, it is almost memoryless.
To obtain an emphasis estimation for a given state using the followon trace, 
we have to simulate a trajectory long enough to go into the mixing stage and visit that particular state,
which is typically difficult in offline training.
This lack of memory is also a cause of the large approximation error. 

In this paper, we propose a novel stochastic approximation algorithm,
Gradient Emphasis Learning (GEM),
to learn $m_\pi$ using function approximation.
GEM can track the true emphasis $m_\pi$ under a changing target policy $\pi$.

\textbf{Algorithm Design: }
We consider linear function approximation, and 
our estimate for $m_\pi$ is $m \doteq Xw$, where $w \in \mathbb{R}^{K_1}$ is the learnable parameters.
For a vector $y \in \mathbb{R}^{\ns}$, 
we define an operator $\mop$ as
$\mop y \doteq i + \gamma D^{-1}P_\pi^\top D y$.
\begin{proposition}
\label{prop:tabular_mop}
$\mop$ is a contraction mapping w.r.t.\ some
weighted maximum norm and $m_\pi$ is its unique fixed point.
\end{proposition}
The proof involves arguments from \citet{bertsekas1989parallel}, where the choice of the weighted maximum norm depends on $\gamma D^{-1}P_\pi^\top D$.
Our operator $\mop$ is a generalization of the discounted COP-TD operator $Y_{\hat{\gamma}}$ \citep{gelada2019off}, 
where $Y_{\hat{\gamma}}y \doteq (1 - \hat{\gamma})1 + \hat{\gamma}D^{-1}P_\pi^\top Dy$ 
and $\hat{\gamma}$ is a scalar similar to $\gamma$.
They show that $Y_{\hat{\gamma}}$ is contractive only when $\hat{\gamma}$ is small enough.
Here our Proposition~\ref{prop:tabular_mop} proves contraction for any $\gamma < 1$.
Although $\mop$ and $Y_{\hat{\gamma}}$ are similar, they are designed for different purposes. 
Namely, $Y_{\hat{\gamma}}$ is designed to learn a density ratio, while $\mop$ is designed to learn the emphasis.
Emphasis generalizes density ratio in that users are free to choose the interest $i$ in $\mop$.

Given Proposition~\ref{prop:tabular_mop}, it is tempting to compose a semi-gradient update rule for updating $w$ analogously to discounted COP-TD, 
where the incremental update for $w_t$ is $(i(S_{t+1}) + \gamma \rho_t x_t^\top w_t - x_{t+1}^\top w_t) x_{t+1}$.
This semi-gradient update, however, can diverge for the same reason as the divergence of off-policy linear TD:
the key matrix $D(I - \gamma P_\pi)$ is not guaranteed to be negative semi-definite (see \citet{sutton2016emphatic}). 
Motivated by GTD methods, 
we seek an approximate solution $m$ that satisfies $m = \Pi \mop m$ via minimizing a projected objective
$||\Pi \bar{\delta}_w ||_D^2$,
where $\bar{\delta}_w \doteq \mop (Xw) - Xw$. 
For reasons that will soon be clear, we also include ridge regularization, yielding the objective
\begin{align}
J^{m_\pi}(w) \doteq \textstyle{\frac{1}{2}}||\Pi \bar{\delta}_w ||_D^2 + \textstyle{\frac{1}{2}}\eta || w ||^2,
\label{eq:objreg}
\end{align}
where $\eta > 0$ is the weight of the ridge term.
We can now compute $\nabla_w J^{m_\pi}(w)$ following a similar routine as \citet{sutton2009fast}.
When sampling $\nabla_w J^{m_\pi}(w)$, 
we use another set of parameters $\kappa \in \R^{K_1}$ to address the double sampling issue as proposed by \citet{sutton2009fast}.
See \citet{sutton2009fast} for details of the derivation.
This derivation, however, provides only an intuition behind GEM and has little to do with the actual convergence proof for two reasons.
First, in an actor-critic setting, $\pi$ keeps changing, as does $J^{m_\pi}$. 
Second, we consider sequential Markovian data $\{S_0, A_0, S_1, \dots\}$.
The proof in \citet{sutton2009fast} assumes i.i.d.\ data, i.e., each state $S_t$ is sampled from $d_\mu$ independently.
Compared with the i.i.d.\ assumption, the Markovian assumption is more practical in RL problems.
We now present the GEM algorithm, which updates $\kappa$ and $w$ recursively as
\begin{align}
{\rm \textbf{GEM:}}&\\
    \bar{\delta}_t &\leftarrow i(S_{t+1}) + \gamma \rho_t x_t^\top w_t - x_{t+1}^\top w_t, \\
    \label{eq:gem-v}
    \kappa_{t+1} &\leftarrow \kappa_t + \alpha_t ( \bar{\delta}_t  - x_{t+1}^\top \kappa_t ) x_{t+1}, \\
    \label{eq:gem-w}
    w_{t+1} &\leftarrow w_t + \alpha_t \big( (x_{t+1} - \gamma \rho_t x_t )x_{t+1}^\top \kappa_t - \eta w_t \big),
\end{align}
where $\eta > 0$ is a constant, ${\alpha_t}$ is a deterministic sequence satisfying the Robbins-Monro condition~\citep{robbins1951stochastic}, 
i.e., $\{\alpha_t\}$ is non-increasing positive and 
$\sum_t \alpha_t = \infty, \sum_t \alpha_t^2 < \infty$. 
Similar to \citet{sutton2009fast}, we define $d_t^\top \doteq [\kappa_t^\top, w_t^\top]$ and rewrite the GEM update as
\begin{align}
    d_{t+1} = d_t + \alpha_t (h(Y_t) - G_{\theta_t}(Y_t) d_t),
\end{align}
where $Y_t \doteq (S_t, A_t, S_{t+1})$. 
With $y \doteq (s, a, \p{s})$, we define 
\begin{align}
\label{eq:GEM_G}
A_\theta(y) &\doteq x(s^\prime)(x(s^\prime) - \gamma \rho_\theta(s, a) x(s))^\top, \\
C(y) &\doteq {x}(s^\prime) x(s^\prime)^\top, \\
G_{\theta}(y) &\doteq \left[ {\begin{array}{*{20}{c}}
    { C(y) }&{ A_\theta(y) }\\
    { -A_\theta(y)^\top}& \eta I
    \end{array}} \right],
h(y) \doteq \left[ {\begin{array}{*{20}{c}}
    {i(\p{s})x(\p{s})}\\
    0
    \end{array}} \right].
\end{align}
Let $\dy(y) \doteq d_\mu(s) \mu(a|s) p(s^\prime|s, a)$, the limiting behavior of GEM is then governed by
\begin{align}
A(\theta) &\doteq \mathbb{E}_\dy[A_\theta(y)] = X^\top (I - \gamma P_\theta^\top)D X, \\
\label{eq:GEM_G_bar}
\bar{G}(\theta) &\doteq \E_\dy[G_\theta(y)] = \left[ {\begin{array}{*{20}{c}}
    { C }&{ A(\theta) }\\
    { -A(\theta)^\top}& \bf{\eta I}
    \end{array}} \right], \\
\bar{h} &\doteq \E_\dy[h(y)] = \left[ {\begin{array}{*{20}{c}}
    {X^\top Di}\\
    0
    \end{array}} \right].
\end{align}
Readers familiar with GTD2 \citep{sutton2009fast} may find that the $\bar{G}(\theta)$ in Eq~\eqref{eq:GEM_G_bar} is different from its counterpart in GTD2 in that
the bottom right block of $\bar{G}(\theta)$ is $\eta I$ while that block in GTD2 is $0$. 
This $\eta I$ results from the ridge regularization in the objective $J^{m_\pi}(w)$,
and this block has to be strictly positive definite\footnote{In this paper, by positive definiteness for an asymmetric square matrix $X$, we mean there exists a constant $\epsilon > 0$ such that $\forall y, y^\top X y \geq \epsilon ||y||^2$.} to ensure the positive definiteness of $\bar{G}(\theta)$.
In general, any regularization in the form of $||w_t||_\Xi^2$ is sufficient.
We assume the ridge to simplify notation.

As we consider an actor-critic setting where the policy $\theta$ is changing every step,
we pose the following condition on the changing rate of $\theta$:
\begin{condition}
\label{con:actor_change_rate}
(Assumption 3.1(3) in \citet{konda2002thesis})
The random sequence $\{\theta_t\}$ satisfies $|| \theta_{t+1} - \theta_t|| \leq \beta_t H_t$, where $\{H_t\}$ is some nonnegative process with bounded moments and $\{\beta_t\}$ is a nonincreasing deterministic sequence satisfying the Robbins-Monro condition such that $\sum_t (\frac{\beta_t}{\alpha_t})^d < \infty$ for some $d > 0$.
\end{condition}
When we consider a policy evaluation setting where $\theta$ is fixed, this condition is satisfied automatically.
We show later that this condition is also satisfied in COF-PAC.
We now characterize the asymptotic behavior of GEM. 
\begin{theorem}
\label{prop:convergence_GEM}
    (Convergence of GEM) Under Assumptions (\ref{assum:mdp}, \ref{assu:non_singular}) and Condition~\ref{con:actor_change_rate}, the iterate $\{d_t\}$ generated by \eqref{eq:gem-w} satisfies $\sup_t ||d_t|| < \infty$ and
    $\lim_{t \rightarrow \infty} ||\bar{G}(\theta_t)d_t - \bar{h}|| = 0 $ 
    almost surely. 
\end{theorem}
\begin{lemma}
\label{lem:G_inverse}
Under Assumptions~(\ref{assum:mdp},\ref{assu:non_singular}), when $\eta > 0$,
$\bar{G}(\theta)$ is nonsingular and $\sup_\theta ||\bar{G}(\theta)^{-1}|| < \infty$.
\end{lemma}
By simple block matrix inversion, Theorem~\ref{prop:convergence_GEM} implies 
\begin{align}
\label{eq:gem_convergence}
\textstyle
&\lim_{t\rightarrow \infty} || w^*_{\theta_t}(\eta) - w_t || = 0, \text{where} \\
&w^*_\theta(\eta) \doteq (A(\theta)^\top C^{-1} A(\theta) + \eta I)^{-1}A(\theta)^\top C^{-1} X^\top Di.
\end{align}
\citet{konda2002thesis} provides a general theorem for stochastic approximation algorithms to track a slowly changing linear system. 
To prove Theorem~\ref{prop:convergence_GEM}, we verify that GEM indeed satisfies all the assumptions (listed in the appendix) in Konda's theorem.
Particularly, that theorem requires $\bar{G}(\theta)$ to be strictly positive definite,
which is impossible if $\eta = 0$.
This motivates the introduction of the ridge regularization in $J^{m_\pi}$ defined in Eq.~\eqref{eq:objreg}.
\emph{Namely, the ridge regularization is essential in the convergence of GEM under a slowly changing target policy.}
Introducing regularization in the GTD objective is not new. 
\citet{mahadevan2014proximal} introduce the proximal GTD learning framework to integrate GTD algorithms with first-order optimization-based regularization via saddle-point formulations and proximal operators.
\citet{yu2017convergence} introduces a general regularization term for improving robustness.
\citet{du2017stochastic} introduce ridge regularization to improve the convexity of the objective.
However, their analysis is conducted with the saddle-point formulation of the  GTD objective~\citep{liu2015finite,macua2015distributed}
and requires a fixed target policy, 
which is impractical in our control setting. 
We are the first to establish the tracking ability of GTD-style algorithms under a slowly changing target policy by introducing ridge regularization,
which ensures the driving term $\bar{G}(\theta)$ is strictly positive definite.
Without this ridge regularization, we are not aware of any existing work establishing this tracking ability.
Note our arguments do not apply when $\eta = 0$ and $\pi$ is changing,
which is an open problem.
However, if $\eta=0$ and $\pi$ is fixed, we can use arguments from \citet{yu2017convergence} to prove convergence.
In this scenario, assuming $A(\theta)$ is nonsingular,
$\{w_t\}$ converges to $w_\theta^*(0) = A(\theta)^{-1}X^\top Di$ and we have
\begin{proposition}
\label{prop:linear_mop}
    $\Pi\mop(Xw_\theta^*(0)) = Xw_\theta^*(0)$. 
\end{proposition}

Similarly, we introduce ridge regularization in the $q$-value analogue of GTD2, which we call GQ2. 
GQ2 updates $u$ recursively as
\begin{align}
{\rm \textbf{GQ2:}}&\\
    \delta_t &\leftarrow R_{t+1} + \gamma \rho_{t+1} \tilde{x}_{t+1}^\top u_t - \tilde{x}_t^\top u_t, \\
    \label{eq:gq-v}
    \tilde{\kappa}_{t+1} &\leftarrow \tilde{\kappa}_t + \alpha_t ( \delta_t  - \tilde{x}_t^\top \tilde{\kappa_t} ) \tilde{x}_t, \\
    \label{eq:gq-w}
    u_{t+1} &\leftarrow u_t + \alpha_t \big( (\tilde{x}_t - \gamma \rho_{t+1} \tilde{x}_{t+1} ) \tilde{x}_t^\top \tilde{\kappa}_t - \eta u_t \big).
\end{align}
Similarly, we define $\tilde{d}^\top_t \doteq [\tilde{\kappa}^\top_t, {u}_t^\top]$,
\begin{align}
 \tilde{A}(\theta) &\doteq \tilde{X}^\top \tilde{D} (I - \gamma \tilde{P}_\theta) \tilde{X}, \\
\tilde{G}(\theta) &\doteq \left[ {\begin{array}{*{20}{c}}
    { \tilde{C} }&{ \tilde{A}(\theta) }\\
    { -\tilde{A}(\theta)^\top}& \eta I
    \end{array}} \right],
\tilde{h} \doteq \left[ {\begin{array}{*{20}{c}}
    {\tilde{X}^\top \tilde{D} \tilde{r}}\\
    0
    \end{array}} \right].
\end{align}
\begin{theorem}
\label{prop:convergence_GQ}
    (Convergence of GQ2) Under Assumptions (\ref{assum:mdp}, \ref{assu:non_singular}) and Condition~\ref{con:actor_change_rate}, the iterate $\{\tilde{d}_t\}$ generated by \eqref{eq:gq-w} satisfies $\sup_t || \tilde{d}_t || < \infty$ and
    $\lim_{t \rightarrow \infty} ||\tilde{G}(\theta_t)\tilde{d}_t - \tilde{h}|| = 0$ 
    almost surely.
\end{theorem}
Similarly, we have
\begin{align}
\label{eq:gq_convergence}
\textstyle
&\lim_{t\rightarrow \infty} || u^*_{\theta_t}(\eta) - u_t || = 0, \text{where} \\
&u^*_\theta(\eta) \doteq (\tilde{A}(\theta)^\top \tilde{C}^{-1} \tilde{A}(\theta) + \eta I)^{-1}\tilde{A}(\theta)^\top \tilde{C}^{-1} \tilde{X}^\top \tilde{D}\tilde{r}.
\end{align}
Comparing the update rules of GEM and GQ2, 
it now becomes clear that GEM is ``reversed'' GQ2.
In particular, the $A(\theta)$ in GEM is the ``transpose'' of the $\tilde{A}(\theta)$ in GQ2. 
Such reversed TD methods have been explored by 
\citet{hallak2017consistent,gelada2019off},
both of which rely on the operator $D^{-1}P_\pi^\top D$ introduced by \citet{hallak2017consistent}.
Previous methods implement this operator 
under the semi-gradient paradigm \citep{sutton1988learning}.
By contrast, GEM is a full gradient.
The techniques in GEM can be applied immediately to the discounted COP-TD \citep{gelada2019off} to improve its convergence from a small enough $\hat{\gamma}$ to any $\hat{\gamma} < 1$.
Applying GEM-style update to COP-TD \citep{hallak2017consistent} is still an open problem as COP-TD involves a nonlinear projection, 
whose gradient is hard to compute.

\section{Convergent Off-Policy Actor-Critic}
To estimate $\nabla J(\theta)$, we use GEM and GQ2 to estimate $m_\pi$ and $q_\pi$ respectively, 
yielding COF-PAC (Algorithm~\ref{alg:cof-pac}).
In COF-PAC, we require both $\{\alpha_t\}$ and $\{\beta_t\}$ to be deterministic and nonincreasing and satisfy the Robbins-Monro condition.
Furthermore, there exists some $d > 0$ 
such that $\sum_t (\frac{\beta_t}{\alpha_t})^d < \infty$.
These are common stepsize conditions in two-timescale algorithms (see \citet{borkar2009stochastic}).
Like \citet{konda2002thesis}, we also use adaptive stepsizes $\Gamma_1: \R^{K_1} \rightarrow \R$ and $\Gamma_2: \R^{K_2} \rightarrow \R$ to ensure $\theta$ changes slowly enough. 
We now pose the same condition on $\Gamma_i (i=1,2)$ as \citet{konda2002thesis}:
(1) $|\Gamma_i(d)| < \infty$ 
(2) $|\Gamma_i(d)| \, ||d|| < \infty$
(3) For all $||d|| < C_1$, there exists $C_2 > 0$ such that $|\Gamma_i(d)| \geq C_2$
(4) $\Gamma_i$ is Lipschitz continuous. 
\citet{konda2002thesis} provides an example for $\Gamma_i$.
Let $C_0 > 0$ be some constant, then we define $\Gamma_i$ as
$\Gamma_i(d) = \mathbb{I}_{||d|| < C_0} + \textstyle{\frac{1 + C_0}{1 + ||d||}} \mathbb{I}_{||d|| \geq C_0}$, 
where $\mathbb{I}$ is the indicator function.
It is easy to verify that the above conditions on stepsizes ($\alpha_t, \beta_t, \Gamma_1, \Gamma_2$), 
together with Assumptions~(\ref{assum:mdp},\ref{assu:policy_params}), 
ensure that $\Gamma_1(w_t)\Gamma_2(u_t)\Delta_t$ is bounded.
Condition~\ref{con:actor_change_rate} on the policy changing rate, therefore, indeed holds.
Consequently, Theorems~\ref{prop:convergence_GEM} and~\ref{prop:convergence_GQ} hold when the target policy $\pi$ is updated according to COF-PAC.
\begin{algorithm}
\caption{COF-PAC}
\label{alg:cof-pac}
\begin{algorithmic}
\ENSURE $\eta > 0$
\STATE Initialize $w_0, \kappa_0, u_0, \tilde{\kappa}_0, \theta_0$
\STATE $t \gets 0$
\STATE Get $S_0, A_0$
\WHILE{True}
\STATE Execute $A_t$, get $R_{t+1}, S_{t+1}$
\STATE Sample $A_{t+1} \sim \mu(\cdot | S_{t+1})$
\STATE $\bar{\delta}_t \gets i(S_{t+1}) + \gamma \rho_t x_t^\top w_t - x_{t+1}^\top w_t$
\STATE $\kappa_{t+1} \gets \kappa_t + \alpha_t ( \bar{\delta}_t  - x_{t+1}^\top \kappa_t ) x_{t+1}$
\STATE $w_{t+1} \gets w_t + \alpha_t \big( (x_{t+1} - \gamma \rho_t x_t )x_{t+1}^\top \kappa_t - \eta w_t \big)$
\STATE $\delta_t \gets R_{t+1} + \gamma \rho_{t+1} \tilde{x}_{t+1}^\top u_t - \tilde{x}_t^\top u_t$
\STATE $\tilde{\kappa}_{t+1} \gets \tilde{\kappa}_t + \alpha_t ( \delta_t  - \tilde{x}_t^\top \tilde{\kappa_t} ) \tilde{x}_t$
\STATE $u_{t+1} \gets u_t + \alpha_t \big( (\tilde{x}_t - \gamma \rho_{t+1} \tilde{x}_{t+1} ) \tilde{x}_t^\top \tilde{\kappa}_t - \eta u_t \big)$
\STATE $\Delta_t \gets \rho_t (w_t^\top x_t) (u_t^\top \tilde{x}_t) \nabla \log \pi_\theta(A_t|S_t)$
\STATE $\theta_{t+1} \gets \theta_t + \beta_t \Gamma_1(w_t) \Gamma_2(u_t)  \Delta_t $ 
\STATE $t \gets t + 1$ 
\ENDWHILE 
\end{algorithmic}
\end{algorithm}

We now characterize the asymptotic behavior of COF-PAC. 
The limiting policy update in COF-PAC is
\begin{align}
\textstyle
\hat{g}(\theta) \doteq \sum_s d_\mu(s) &\big(x(s)^\top w^*_\theta(\eta) \big) \textstyle{\sum_a} \\
&\mu(a|s) \psi_\theta(s, a) \big( \tilde{x}(s, a)^\top u^*_\theta(\eta) \big).
\end{align}
The bias introduced by the estimates $m$ and $q$ is 
$b(\theta) \doteq \nabla J(\theta) - \hat{g}(\theta)$,
which determines the asymptotic behavior of COF-PAC:
\begin{theorem}
    \label{thm:cof-pac}
    (Convergence of COF-PAC) Under Assumptions (\ref{assum:mdp}-\ref{assu:policy_params}), the iterate $\{\theta_t\}$ generated by COF-PAC (Algorithm~\ref{alg:cof-pac}) satisfies
        $\lim \textstyle{\inf_t} \Big( ||\nabla J(\theta_t)|| - ||b(\theta_t)|| \Big) \leq 0$,
    almost surely, i.e., $\{\theta_t\}$ visits any neighborhood of the set
    $\{\theta : ||\nabla J(\theta)|| \leq ||b(\theta)|| \}$
    infinitely many times.
\end{theorem} 
The proof is inspired by \citet{konda2002thesis}.
According to Theorem~\ref{thm:cof-pac}, 
COF-PAC reaches the same convergence level as the canonical on-policy actor-critic \citep{konda2002thesis}. 
Together with the fact that $\nabla J(\theta)$ is Lipschitz continuous and $\beta_t$ is diminishing,
it is easy to see $\theta_t$ will eventually remain in the neighborhood $\{\theta : ||\nabla J(\theta)|| \leq ||b(\theta)|| \}$ in Theorem~\ref{thm:cof-pac} for arbitrarily long time.
When $\pi_\theta$ is close to $\mu$ in the sense
of the following Assumption~\ref{assum:bound_of_bias}(a),
we can provide an explicit bound for the bias $b(\theta)$.
However, failing to satisfy Assumption~\ref{assum:bound_of_bias} does not necessarily imply the bias is large. 
The bound here is indeed loose and is mainly
to provide an intuition for the source of the bias.

\begin{assumption}
\label{assum:bound_of_bias}
(a) The following two matrices are positive semidefinite :
\begin{align}
F_\theta &\doteq \left[ {\begin{array}{*{20}{c}}
    { C }&{ X^\top P_\theta^\top DX }\\
    { X^\top DP_\theta X}& C 
    \end{array}} \right], \\
\tilde{F}_\theta &\doteq \left[ {\begin{array}{*{20}{c}}
    { \tilde{C} }&{ \tilde{X}^\top \tilde{D} \tilde{P}_\theta \tilde{X} }\\
    { \tilde{X}^\top \tilde{P}^\top_\theta \tilde{D} \tilde{X}}& \tilde{C} 
    \end{array}} \right].
\end{align}
(b) $\inf_\theta |\det(A(\theta))| > 0, \inf_\theta |\det(\tilde{A}(\theta))| > 0$. \\
(c) The Markov chain induced by $\pi_\theta$ is ergodic.
\end{assumption}
\begin{remark}
Part (a) is from \citet{kolter2011fixed}, which ensures $\pi_\theta$ is not too far away from $\mu$. 
The non-singularity of $A(\theta)$ and $\tilde{A}(\theta)$ for each fixed $\theta$ is commonly assumed \citep{sutton2009fast,sutton2009convergent,maei2011gradient}.
In part (b), we make a slightly stronger assumption that their determinants do not approach 0 during the optimization of $\theta$.
\end{remark}

\begin{proposition}
\label{prop:b_theta_bound}
Under Assumptions~(\ref{assum:mdp}-\ref{assum:bound_of_bias}), 
let $d_\theta$ be the stationary distribution under $\pi_\theta$ and define $\tilde{d}_\theta(s, a) \doteq d_\theta(s) \pi_\theta(a|s), D_\theta\doteq diag(d_\theta), \tilde{D}_\theta \doteq diag(\tilde{d}_\theta)$,
we have
\begin{align}
||b(\theta)||_D \leq C_0\eta
&+ C_1 \textstyle{\frac{1 + \gamma \kappa(D^{-\frac{1}{2}} D_\theta^{\frac{1}{2}})}{1-\gamma}}
||m_{\pi_\theta} - \Pi m_{\pi_\theta}||_D \\
&+ C_2\textstyle{\frac{1 + \gamma \kappa(\tilde{D}^{-\frac{1}{2}} \tilde{D}_\theta^{\frac{1}{2}})}{1-\gamma}} ||q_{\pi_\theta} - \tilde{\Pi} q_{\pi_\theta}||_{\tilde{D}},
\end{align}
where $\kappa(\cdot)$ is the condition number of a matrix w.r.t.\ $\ell_2$ norm and $C_0, C_1, C_2$ are some positive constants.
\end{proposition}
The bias $b(\theta)$ comes from the bias of both the $q_\pi$ estimate and the $m_\pi$ estimate.
The bound of the $q_\pi$ estimate follows directly from \citet{kolter2011fixed}.
The proof from \citet{kolter2011fixed}, however, can not be applied to analyze the $m_\pi$ estimate until Lemma~\ref{lem:p_pi_norm} is established.

\textbf{Compatible Features: }
One possible approach to eliminate the bias $b(\theta)$
is to consider \emph{compatible features} as in the canonical on-policy actor-critic \citep{sutton2000policy,konda2002thesis}.
Let $\Psi$ be a subspace and $<\cdot, \cdot>_\Psi$ be an inner product, 
which induces a norm $||\cdot||^\Psi$.
We define a projection $\Pi_\Psi$ as $\Pi_\Psi y \doteq \arg\min_{\bar{y} \in \Psi}||\bar{y} - y||^\Psi$.
For any vector $y$ and a vector $\bar{y} \in \Psi$, we have $<y - \Pi_\Psi y, \bar{y}>_\Psi = 0$ by Pythagoras.
Based on this equality, \citet{konda2002thesis} designs compatible features for an on-policy actor-critic.
Inspired by \citet{konda2002thesis}, we now design compatible features for COF-PAC.

Let $\hat{m}_\theta, \hat{q}_\theta$ be estimates for $m_\pi, q_\pi$.
With slight abuse of notations, we define 
\begin{align}
\hat{g}(\theta) \doteq \textstyle{\sum_s} d_\mu(s)\hat{m}_\theta(s) \textstyle{\sum_a} \mu(a|s)\psi_\theta(s, a)\hat{q}_\theta(s, a),
\end{align}
which is the limiting policy update.
The bias $\nabla J(\theta) - \hat{g}(\theta)$ can then be decomposed as $b_1(\theta) + b_2(\theta)$, where
\begin{align}
&b_1(\theta) \doteq \textstyle{\sum_s} d_\mu(s) (m_\pi(s) - \hat{m}_\theta(s)) \phi_1^\theta(s), \\
&\phi_1^\theta(s) \doteq \textstyle{\sum_a} \mu(a|s)\psi_\theta(s, a) \hat{q}_\theta(s, a), \\
&b_2(\theta) \doteq \textstyle{\sum_{s, a}} d_{\mu, m}(s, a) \phi_2^\theta(s, a)(q_\pi(s, a) - \hat{q}_\theta(s, a)), \\
&d_{\mu, m}(s, a) \doteq d_\mu(s) m_\pi(s) \mu(a | s), \, \phi_2^\theta(s, a) \doteq \psi_\theta(s, a).
\end{align}
For an $i \in [1, \dots, K]$, we consider $\phi_{1,i}^\theta \in \R^{|\mathcal{S}|}$, where $\phi_{1,i}^\theta(s)$ is the $i$-th element of $\phi_1^\theta(s) \in \R^K$.
Let $\Psi_1$ denote the subspace in $\R^{|\mathcal{S}|}$ spanned by $\{\phi_{1, i}^\theta\}_{i=1,\dots, K}$.
We define an inner product $<y_1, y_2>_{\Psi_1} \doteq \sum_s d_\mu(s)y_1(s)y_2(s)$.
Then we can write $b_{1, i}(\theta)$, the $i$-the element of $b_1(\theta)$, as
\begin{align}
b_{1, i}(\theta) = <m_\pi - \hat{m}_\theta, \phi_{1, i}^\theta>_{\Psi_1}.
\end{align}
If our estimate $\hat{m}_\theta$ satisfies $\hat{m}_\theta = \Pi_{\Psi_1} m_\pi$, 
we have $b_1(\theta) = 0$.
This motivates learning the estimate $\hat{m}_\theta$ via minimizing $J_{\Psi_1} \doteq ||\Pi_{\Psi_1}m_\pi - \hat{m}_\theta||^{\Psi_1}$.
One possibility is to consider linear function approximation for $\hat{m}_\theta$ and use $\{\phi_{1, i}^\theta\}$ as features.
Similarly, we consider the subspace $\Psi_2$ in $\R^{N_{sa}}$ 
spanned by $\{\phi_{2, i}^\theta\}$
and define the inner product according to $d_{\mu, m}$.
We then aim to learn $\hat{q}_\theta$ via minimizing $J_{\Psi_2} \doteq ||\Pi_{\Psi_2} q_\pi - \hat{q}_\theta ||^{\Psi_2}$.
Again, we can consider linear function approximation with features $\{\phi_{2, i}^\theta\}$.
In general, any feature, whose feature space contains $\Psi_1$ or $\Psi_2$, are compatible features.
Due to the change of $\theta$, compatible features usually change every time step \citep{konda2002thesis}.
Note if we consider a state value critic instead of a state-action value critic, 
the computation of compatible features will involve the transition kernel $p$, 
to which we do not have access.

In the on-policy setting, 
Monte Carlo or TD(1) can be used to train a critic with compatible features \citep{sutton2000policy,konda2002thesis}.
In the off-policy setting,
one could consider a GEM analogue of GTD($\lambda$) \citep{yu2017convergence} with $\lambda=1$ to minimize $J_{\Psi_1}$.
To minimize $J_{\Psi_2}$, one could consider a $q$-value analogue of ETD($\lambda$) \citep{yu2015convergence} with $\lambda=1$.
We leave the convergent analysis for those analogues under a changing target policy for future work.

\section{Experiments}
We design experiments to answer the following questions: (a) Can GEM approximate the emphasis as promised? (b) Can the GEM-learned emphasis boost performance compared with the followon trace?
All curves are averaged over 30 independent runs. 
Shadowed regions indicate one standard derivation.
All the implementations are made publicly available for future research.\footnote{\url{https://github.com/ShangtongZhang/DeepRL}}

\begin{figure}[h]
    \center
    \includegraphics[width=0.35\textwidth]{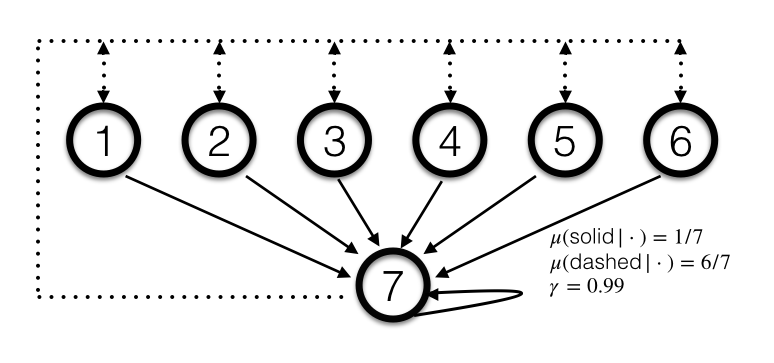}
    \caption{\label{fig:baird} A variant of Baird's counterexample. This figure is adapted from \citet{sutton2018reinforcement}. The \texttt{solid} action always leads to the state 7 and a reward 0, and the \texttt{dashed} action leads to states 1 - 6 with equal probability and a reward $+1$.
    }
\end{figure}

\textbf{Approximating Emphasis:}
We consider variants of Baird's counterexample \citep{baird1995residual,sutton2018reinforcement} as shown in Figure~\ref{fig:baird}.
There are two actions and the behavior policy $\mu$ always chooses the \texttt{dashed} action with probability $\frac{6}{7}$.
The initial state is chosen from all the states with equal probability, and the interest $i$ is 1 for all states.
We consider four different sets of features: original features, one-hot features, zero-hot features, and aliased features. 
Original features are the features used by \citet{sutton2018reinforcement}, 
where the feature for each state lies in $\mathbb{R}^8$ 
(details in the appendix).
This set of features is uncommon 
as in practice the number of states is usually much larger than the number of features.
One-hot features use one-hot encoding,
where each feature lies in $\mathbb{R}^7$,
which indeed degenerates to a tabular setting.
Zero-hot features are the complements of one-hot features, 
e.g., the feature of the state 1 is $[0, 1, 1, 1 ,1 ,1, 1]^\top \in \mathbb{R}^7$.
The quantities of interest, e.g., $m_\pi$ and $v_\pi$, can be expressed accurately under all the three sets of features.
In the fourth set of features, we consider state aliasing.
Namely, we still consider the original features, but now the feature of the state 7 is modified to be identical as the feature of the state 6.
The last two dimensions of features then become identical for all states, and therefore we removed them, resulting in features lying in $\mathbb{R}^6$.
Now the quantities of interest may not lie in the feature space.

\begin{figure}
  \center
  \includegraphics[width=0.5\textwidth]{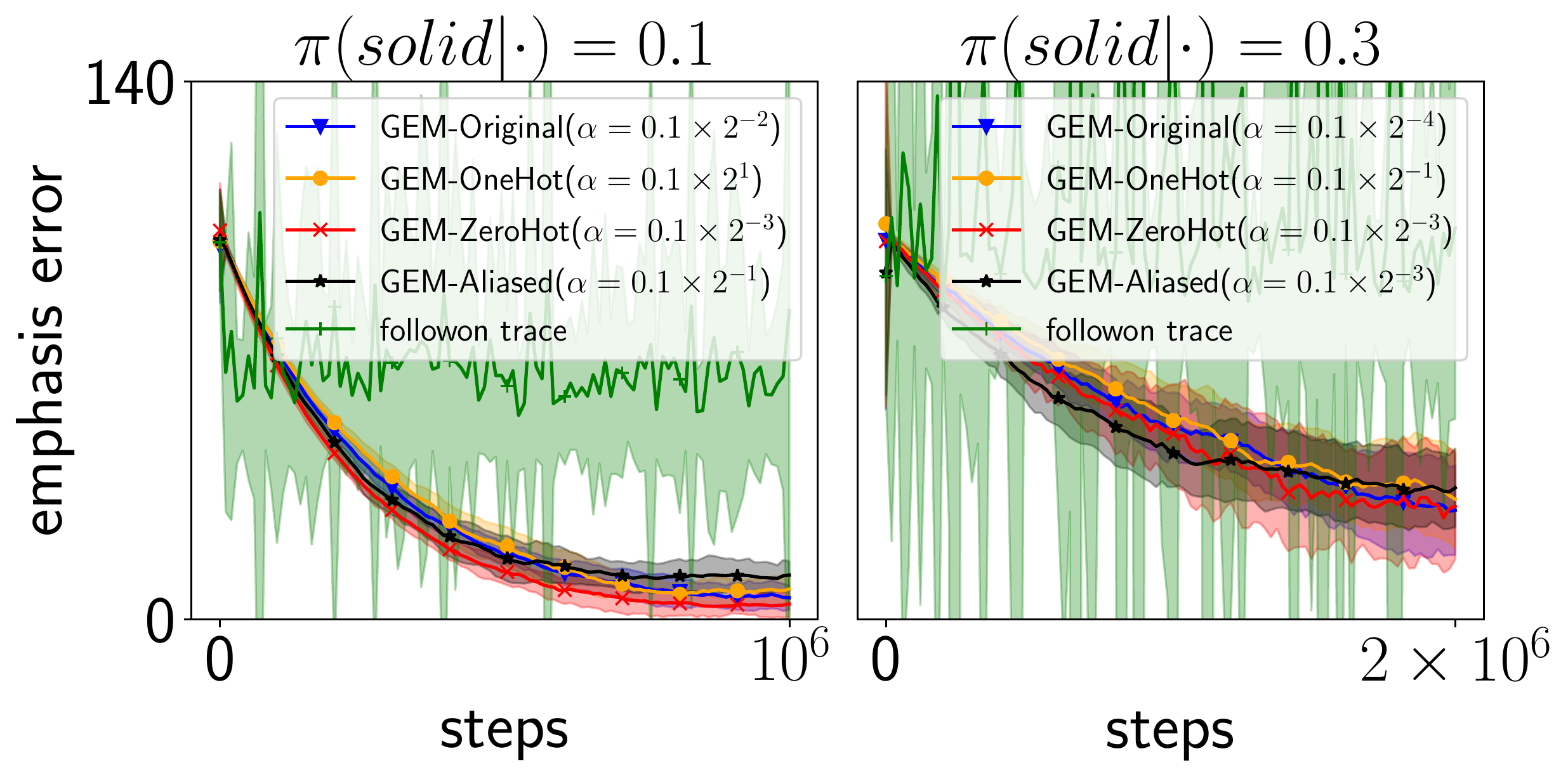}
  \caption{\label{fig:emphasis_error} Averaged emphasis approximation error in last 1000 steps for the followon trace and GEM with different features. Learning rates used are bracketed. }
\end{figure}

In this section, we compare the accuracy of approximating the emphasis $m_\pi$ with GEM (Eq~\eqref{eq:gem-w}) and the followon trace (Eq~\eqref{eq:update_M_t}).
We report the emphasis approximation error in Figure~\ref{fig:emphasis_error}.
At time step $t$, the emphasis approximation error is computed as $|M_t - m_\pi(S_t)|$ and $|w_t^\top x(S_t) - m_\pi(S_t)|$ for the followon trace and GEM respectively, 
where $m_\pi$ is computed analytically,
$M_{-1}=0$, and $w_0$ is drawn from a unit normal distribution.
For GEM, we consider a fixed learning rate $\alpha$ and tune it from $\{0.1 \times 2^1, \dots, 0.1 \times 2^{-6}\}$.
We consider two target policies: $\pi(\texttt{solid}| \cdot) = 0.1$ and $\pi(\texttt{solid}| \cdot) = 0.3$.

\begin{figure*}[t]
  \center
  \includegraphics[width=0.9\textwidth]{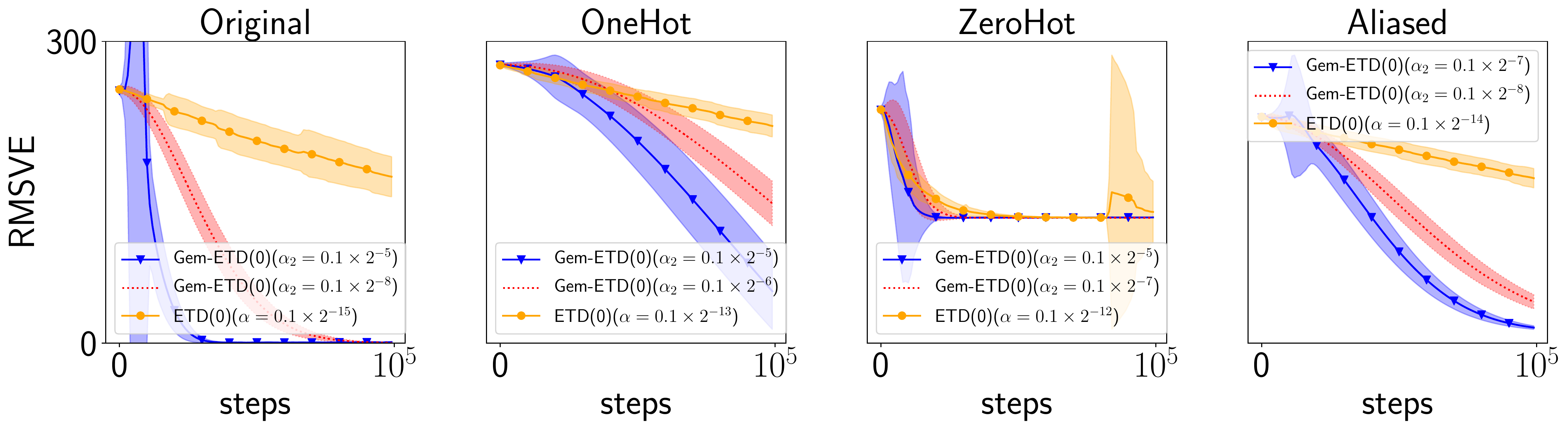}
  \caption{\label{fig:etd_error} 
  Averaged RMSVE in recent 1000 steps for GEM-ETD(0) and ETD(0) with four different sets of features.}
\end{figure*}

As shown in Figure~\ref{fig:emphasis_error}, the GEM approximation enjoys lower variance than the followon trace approximation
and has a lower approximation error under all four sets of features.
Interestingly, when the original features are used, the $C$ matrix is indeed singular, which violates  Assumption~\ref{assu:non_singular}.
However, the algorithm does not diverge.
This may suggest that the Assumption~\ref{assu:non_singular} can be relaxed in practice.

\textbf{Policy Evaluation:}
The followon trace $M_t$ is originally used in ETD to reweight updates (Eq~\eqref{eq:update_M_t} and Eq~\eqref{eq:etd_update}). 
We compare ETD(0) with GEM-ETD(0), which updates $\nu$ as
\begin{align}
\nu_{t+1} \gets \nu_t + \alpha_2 \hat{M}_t \rho_t (R_{t+1} + \gamma x_{t+1}^\top \nu_t - x_t^\top \nu_t)x_t^\top,
\end{align}
where $\hat{M}_t \doteq w_t^\top x_t$ and $w_t$ is updated according to GEM (Eq~\eqref{eq:gem-w}) with a fixed learning rate $\alpha_1$. 
If we assume $m_\pi$ lies in the column space of $X$, 
a convergent analysis of GEM-ETD(0) is straightforward.

We consider a target policy $\pi(\texttt{solid}| \cdot) = 0.05$. 
We report the Root Mean Squared Value Error (\emph{RMSVE}) at each time step during training in Figure~\ref{fig:etd_error}.
RMSVE is computed as $||v - v_\pi||_D$, 
where $v_\pi$ is computed analytically.
For ETD(0), we tune the learning rate $\alpha$ from $\{0.1 \times 2^0, \dots, 0.1 \times 2^{-19}\}$.
For GEM-ETD(0), we set $\alpha_1=0.025$ and tune $\alpha_2$ in the same range as $\alpha$. 
For both algorithms, we report the results with learning rates that minimized the area under the curve (AUC) in the solid lines in Figure~\ref{fig:etd_error}.
In our policy evaluation experiments, GEM-ETD(0) has a clear win over ETD(0) under all four sets of features. 
Note the AUC-minimizing learning rate for ETD(0) is usually several orders smaller than that of GEM-ETD(0),
which explains why ETD(0) curves tend to have smaller variance than GEM-ETD(0) curves.
When we decrease the learning rate of GEM-ETD(0) (as indicated by the red dashed lines in Figure~\ref{fig:etd_error}),
the variance of GEM-ETD(0) can be reduced, and the AUC is still smaller than that of ETD(0).

GEM-ETD is indeed a way to trade off bias and variance.
If the states are heavily aliased, the GEM emphasis estimation may be heavily biased, as will GEM-ETD.
We do not claim that GEM-ETD is always better than ETD.
For example, when we set the target policy to $\pi(\texttt{solid}|\cdot) = 1$,
there was no observable progress for both GEM-ETD(0) and ETD(0) with reasonable computation resources.\footnote{This target policy is problematic for GEM-ETD(0) mainly because the magnitude of $\bar{\delta_t}$ in Eq~\eqref{eq:gem-w} varies dramatically across different states,
which makes the supervised learning of $\kappa$ hard.}
When it comes to the bias-variance trade-off, the optimal choice is usually task-dependent.
Our empirical results suggest GEM-ETD is a promising approach for this trade-off.
ETD(0) is a special case of ETD($\lambda, \beta$) \citep{hallak2016generalized}, 
where $\lambda$ and $\beta$ are used for bias-variance trade-off.
Similarly, we can have GEM-ETD($\lambda, \beta$) by introducing $\lambda$ and $\beta$ to our GEM operator $\mop$ analogously to ETD($\lambda, \beta$).
A comparison between ETD($\lambda, \beta$) and GEM-ETD($\lambda, \beta$) is a possibility for future work.

\begin{figure}
  \center
  \includegraphics[width=0.3\textwidth]{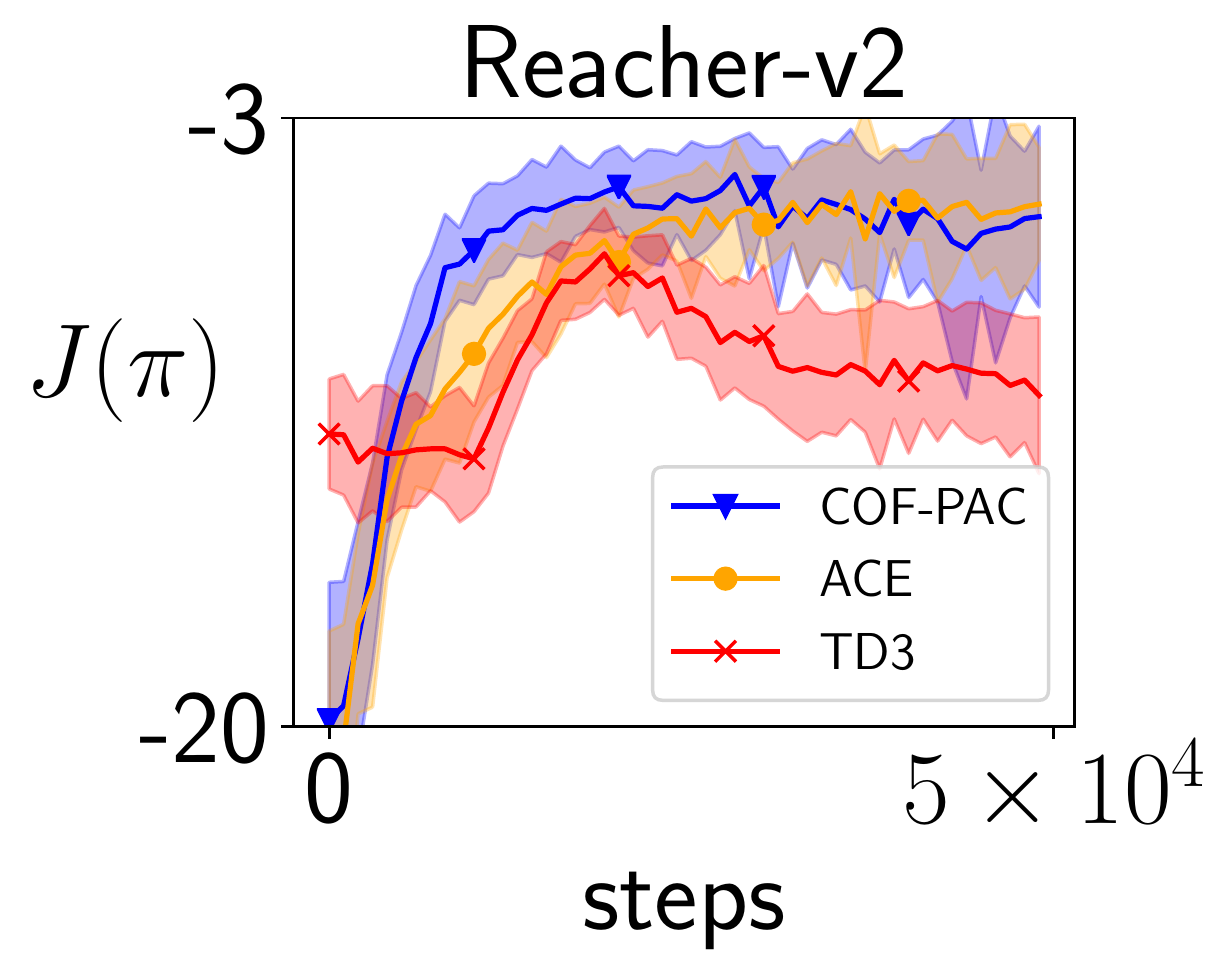}
  \caption{\label{fig:control} 
  Comparison between ACE, COF-PAC and TD3 with a uniformly random behavior policy.}
\end{figure}

\textbf{Control:}
We benchmarked COF-PAC, ACE and TD3 \citep{fujimoto2018addressing} in \texttt{Reacher-v2} from OpenAI Gym~\citep{brockman2016openai}.
Our implementation is based on \citet{zhang2019generalized}, and we inherited their hyperparameters.
Like \citet{gelada2019off,zhang2019generalized}, 
we consider \emph{uniformly random} behavior policy.
Neural networks are used to parameterize $\pi, v_\pi, m_\pi$. 
A semi-gradient version of GEM is used to train $m_\pi$ inspired by the success of semi-gradient methods in large scale RL \citep{mnih2015human}.
Details are provided in the appendix.
We trained both algorithms for $5 \times 10^4$ steps and evaluate $J(\pi)$ every $10^3$ steps.
According to Figure~\ref{fig:control}, COF-PAC solves the task faster than ACE and is more stable than TD3 in this tested domain.

\section{Related Work}
Our work relies on results from~\citet{konda2002thesis} with two fundamental differences:
(1) The work by \citet{konda2002thesis} focuses on only the on-policy setting and cannot be naturally extended to the off-policy counterpart,  while we work on the off-policy setting by incorporating state-of-the-art techniques such as GTD, emphatic learning, and reversed TD.
(2) The learning architecture has substantial differences in that \citet{konda2002thesis} considers one TD critic while we consider two GTD-style critics. 
As the structures of TD algorithms and GTD-style algorithms are dramatically different,
applying Konda's arguments for a TD critic to our two GTD-style critics is not straightforward.

\citet{maei2018convergent} proposes the Gradient Actor-Critic algorithm under a different objective, $\sum_s d_\mu(s) v(s)$, 
for off-policy learning with function approximation.
This objective differs from the excursion objective in that it replaces the true value function $v_\pi$ with an estimate $v$.
Consequently, the optimal policy under this objective depends on the features used to approximate the value function,
and this approximation of the excursion objective can be arbitrarily poor. 
\citet{maei2018convergent} tries to show the convergence of a GTD critic under a slowly changing target policy with results from \citet{konda2002thesis}.
In this paper, we show that GTD has to be regularized before the results from \citet{konda2002thesis} can take over.
Furthermore, the policy gradient estimator \citet{maei2018convergent} proposes is also based on the followon trace.
That estimator tracks the true gradient only in a limiting sense under a fixed $\pi$ \citep[Theorem 2]{maei2018convergent}
and has potentially unbounded variance,
similar to how $M_t$ tracks $m_\pi(S_t)$.
It is unclear if that policy gradient estimator can track the true policy gradient under a changing $\pi$ or not.
To address this issue, we instead use function approximation to learn the emphasis directly.


Off-PAC has inspired the invention of many other off-policy actor-critic algorithms (e.g., off-policy DPG, DDPG, ACER, off-policy EPG, TD3, IMPALA), 
all of which, like Off-PAC, ignore the emphasis
and thus are not theoretically justified under function approximation.
Another line of policy-based off-policy algorithms involves 
reward shaping via policy entropy.
In particular,
SBEED \citep{dai2017sbeed} is an off-policy actor-critic algorithm with a finite-sample analysis on the statistical error.
The convergence analysis of SBEED (Theorem 5 in \citet{dai2017sbeed}) is conducted within a bi-level optimization framework,
assuming the exact solution of the inner optimization problem can be obtained.
With function approximation,
requiring the exact solution is usually impractical due to representation error.
Even if the exact solution is representable,
computing it explicitly is still expensive (c.f. solving a least-squares regression problem with a large feature matrix exactly).
By contrast, our work adopts a two-timescale perspective, 
where we do not need to obtain the exact minimizer of $J^{m_\pi}(w)$ every step.
Other works in this line of research include \citet{nachum2017bridging,o2016combining,schulman2017equivalence,nachum2017trust,haarnoja2017reinforcement,haarnoja2018soft},
which are mainly developed in a tabular setting and do not have a convergence analysis under function approximation.

\citet{liu2019off} propose to reweight the Off-PAC update via the density ratio between $\pi$ and $\mu$.
This density ratio can be learned by either \citet{liu2018breaking} as \citet{liu2019off} did or (discounted) COP-TD, \citet{nachum2019dualdice}, \citet{uehara2019minimax}, \citet{zhang2020gendice}, and GradientDICE \citep{zhang2020gradientdice}. 
The convergence of those density ratio learning algorithms under a slowly changing target policy is, however, unclear.
For GradientDICE with linear function approximation,
it is possible to employ our arguments for proving Theorem~\ref{prop:convergence_GEM} to prove its convergence under a slowly changing target policy
and thus give a convergent analysis for this reweighted Off-PAC in a two-timescale form.
We leave this for future work.
\citet{zhang2019generalized} propose a new objective based on the density ratio from \citet{gelada2019off}, 
yielding Generalized Off-Policy Actor-Critic (Geoff-PAC), 
whose convergence is also unclear.

In concurrent work (AlgaeDICE), \citet{nachum2019algaedice} propose a new objective for off-policy actor-critic and reformulate the policy optimization problem into a minimax problem.
Primal-dual algorithms can then take over.
\citet{nachum2019algaedice} show the primal variable works similarly to an actor, and the dual variable works similarly to a critic.
It is possible to provide a two-timescale convergent analysis for AlgaeDICE when the dual variable is linear and the primal variable is nonlinear using arguments from this paper,
which we also leave for future work.

Although the actor runs at a slower timescale and the critic runs at a faster timescale, 
we remark that applying the two-timescale convergent arguments from Chapter 6.1 of \citet{borkar2009stochastic} to the actor-critic setting with function approximation is in general hard.
Due to function approximation, 
the equilibrium of the ordinary differential equation (ODE) associated with the critic usually depends on features. 
Consequently, the ODE associated with the actor depends on the features as well,
making it hard to analyze.
To eliminate the influence of the representation error resulting from function approximation,
compatible features and eligibility trace with $\lambda = 1$ seem necessary,
both of which, however, do not fit well into arguments from \citet{borkar2009stochastic}.

\section{Conclusion}
We have presented the first provably convergent two-timescale off-policy actor-critic with function approximation via introducing the emphasis critic and establishing the tracking ability of GTD-style algorithms under a slowly changing target policy.
Future work can extend COF-PAC with non-linear critics via considering projection onto the tangent plane as \citet{maei2011gradient}.
Conducting a finite sample analysis of COF-PAC with techniques in \citet{zou2019finite,xu2020non,xu2020improving} is also a possibility for future work.

\section*{Acknowledgments}
SZ is generously funded by the Engineering and Physical Sciences Research Council (EPSRC). This project has received funding from the European Research Council under the European Union's Horizon 2020 research and innovation programme (grant agreement number 637713). The experiments were made possible by a generous equipment grant from NVIDIA. 
BL’s research is funded by the National Science Foundation (NSF) under grant NSF IIS1910794, Amazon Research Award, and Adobe gift fund.

\bibliography{ref}
\bibliographystyle{icml2020}

\onecolumn
\newpage
\appendix

\section{Proofs}
\subsection{Proof of Lemma~\ref{lem:boundess_policy_params}}
\textbf{Lemma~\ref{lem:boundess_policy_params}.} 
\emph{Under Assumptions (\ref{assum:mdp}, \ref{assu:policy_params}), there exists a constant $C_1 < \infty$ such that $\forall (\theta, \bar{\theta})$}
    \begin{align}
    &||\nabla J(\theta)|| \leq C_1,  \\
    &||\nabla J(\theta) - \nabla J(\bar{\theta})|| \leq C_1||\theta - \bar{\theta}||, \\
    &||\textstyle{\frac{\partial^2 J(\theta)}{\partial \theta_i \partial \theta_j}}|| \leq C_1.
    \end{align}
\begin{proof}
We use $||\cdot||_\infty$ to denote infinity norm.
It is a well-known fact that $||P_\theta||_\infty = 1$.
For any $y$, either a vector or a square matrix, we have
\begin{align}
||(I - \gamma P_\theta)^{-1}y||_\infty = ||\sum_{i=0}^{\infty} \gamma^i P_\theta^i y||_\infty \leq \sum_{i=0}^{\infty} \gamma^i ||P_\theta||^i_\infty ||y||_\infty = \frac{||y||_\infty}{1 - \gamma},
\end{align}
implying
\begin{align}
\label{eq:series_bound}
||(I - \gamma P_\theta)^{-1}y|| \leq \frac{\sqrt{|\mathcal{S}|}||y||_\infty}{1 - \gamma}.
\end{align}
(i) Applying Eq~\eqref{eq:series_bound} to $m_\pi$ and $v_\pi$ yields
\begin{align}
||m_\pi|| &= ||D^{-1}(I - \gamma P_\theta^\top)^{-1}Di|| \leq ||D^{-1}(I - \gamma P_\theta^\top)^{-1}||\, || Di || \\
&=||(I-\gamma P_\theta)^{-1}D^{-1}|| \, ||Di||  \leq ||Di|| \frac{\sqrt{|\mathcal{S}|}||D^{-1}||_\infty}{1 - \gamma}, \\
||v_{\pi}|| &= ||(I - \gamma P_{\theta})^{-1}r_{\pi}|| \leq \frac{\sqrt{|\mathcal{S}|} \max_{s, a, \p{s}}r(s, a, \p{s}) }{1 - \gamma}.
\end{align}
According to the analytical expression of $\nabla_\theta J(\theta)$ in  Eq~\eqref{eq:grad_J_theta}, 
it follows easily that $\sup_\theta ||\nabla_\theta J(\theta)|| < \infty$.

(ii) For the sake of clarity, in this part, use $\nabla_\theta$ to denote the gradient w.r.t. one dimension of $\theta$.
We first show $\nabla_\theta v_\pi(s)$ is bounded.
As $v_\pi = r_\pi + \gamma P_\pi v_\pi$, 
we have 
\begin{align}
    \nabla_\theta v_\pi &= \nabla_\theta r_\pi + \gamma P_\pi \nabla_\theta v_\pi + \gamma \nabla_\theta P_\pi v_\pi, \\
    \nabla_\theta v_\pi &= (I - \gamma P_\pi)^{-1}(\nabla_\theta r_\pi + \gamma \nabla_\theta P_\pi v_\pi).
\end{align}
According to Assumptions (\ref{assum:mdp}, \ref{assu:policy_params}), $\sup_\theta ||\nabla_\theta r_\pi + \gamma \nabla_\theta P_\pi v_\pi || < \infty $, Eq~\eqref{eq:series_bound} then implies $\sup_\theta || \nabla_\theta v_\pi || < \infty$.

We then show $\nabla_\theta m_\pi(s)$ is bounded. We have
\begin{align}
    \label{eq:m_fixed_point}
    i + \gamma D^{-1} P_\pi^\top D m_\pi &= i + \gamma D^{-1} P_\pi^\top (I - \gamma P_\pi^\top)^{-1}Di \\
    & = \Big(D^{-1}(I - \gamma P_\pi^\top) +  \gamma D^{-1}P_\pi^\top \Big) (I - \gamma P_\pi^\top)^{-1}Di \\
    & = D^{-1}(I - \gamma P_\pi^\top)^{-1} Di = m_\pi.
\end{align}
Taking gradients in both sides,
\begin{align}
    \nabla_\theta m_\pi &= \gamma D^{-1} \nabla_\theta P_\pi^\top D m_\pi + \gamma D^{-1} P_\pi^\top D \nabla_\theta m_\pi \\
    \nabla_\theta m_\pi &= (I - \gamma D^{-1} P_\pi^\top D)^{-1} \gamma D^{-1} \nabla_\theta P_\pi^\top D m_\pi \\
    &=\Big( D^{-1}(I - \gamma P_\pi^\top) D \Big)^{-1} \gamma D^{-1} \nabla_\theta P_\pi^\top D m_\pi \\
    &= D^{-1}(I - \gamma P_\pi^\top)^{-1} D \gamma D^{-1} \nabla_\theta P_\pi^\top D m_\pi \\
    &= \gamma D^{-1}(I - \gamma P_\pi^\top)^{-1} \nabla_\theta P_\pi^\top D m_\pi, \\
    ||\nabla_\theta m_\pi || &\leq \gamma ||\nabla_\theta P_\pi^\top D m_\pi ||\, ||(I - \gamma P_\pi)^{-1}D^{-1} ||
\end{align}
Eq~\eqref{eq:series_bound} then implies $\sup_\theta ||\nabla_\theta m_\pi|| < \infty$.
We now take gradients w.r.t. $\theta$ in both sides of Eq~\eqref{eq:grad_J_theta} and use the product rule of calculus, 
it follows easily that 
$\sup_\theta ||\frac{\partial^2 J(\theta)}{\partial \theta_i \theta_j} || < \infty$.

(iii)
The bounded Hessian of $J(\theta)$ in (ii) implies $\nabla_\theta J(\theta)$ is Lipschitz continuous.




\end{proof}

\subsection{Proof of Lemma~\ref{lem:p_pi_norm}}
\textbf{Lemma~\ref{lem:p_pi_norm}.}
\emph{Under Assumption~\ref{assum:mdp},
$||P_\pi||_D = ||D^{-1}P_\pi^\top D||_D$}
\begin{proof}
This proof is inspired by \citet{kolter2011fixed}.
\begin{align}
||P_\pi||_D &= \sup_{||x||_D=1} ||P_\pi x||_D = \sup_{||x||_D=1} \sqrt{x^\top P_\pi^\top D P_\pi x} \\
&= \sup_{||y|| = 1} \sqrt{y^\top D^{-\frac{1}{2}} P_\pi^\top D P_\pi D^{-\frac{1}{2}} y} = ||D^{\frac{1}{2}} P_\pi D^{-\frac{1}{2}} || \\
||D^{-1} P_\pi^\top D||_D &= \sup_{||y|| = 1} \sqrt{y^\top D^{-\frac{1}{2}} D P_\pi^\top D^{-1} D D^{-1} P_\pi^\top D D^{-\frac{1}{2}} y} = ||D^{-\frac{1}{2}} P_\pi^\top D^{\frac{1}{2}} ||
\end{align}
The rest follows from the well-known fact that $\ell_2$ matrix norm is invariant under matrix transpose.
\end{proof}

\subsection{Proof of Lemma~\ref{lem:G_inverse}}
\textbf{Lemma~\ref{lem:G_inverse}.}
\emph{Under Assumptions~(\ref{assum:mdp},\ref{assu:non_singular}), when $\eta > 0$,
$\bar{G}(\theta)$ is nonsingular and $\sup_\theta ||\bar{G}(\theta)^{-1}|| < \infty$.}
\begin{proof}
By rule of block matrix determinant, we have
\begin{align}
\det(\bar{G}(\theta)) &= \det(\eta I) \det(C + \frac{1}{\eta} A(\theta)A(\theta)^\top) \\
&\geq \det(\eta I) [\det(C) + \det(\frac{1}{\eta} A(\theta)A(\theta)^\top)] \\
&\geq \det(\eta I) \det(C),
\end{align}
where both inequalities result from the positive semi-definiteness of $A(\theta)A(\theta)^\top$ 
and Assumption~\ref{assu:non_singular}.
$\bar{G}(\theta)$ is therefore nonsingular. 
Recall 
\begin{align}
\bar{G}(\theta)^{-1} = \frac{\text{adj}(\bar{G}(\theta))}{\det(\bar{G}(\theta))},
\end{align}
where adj($\bar{G}(\theta)$) is the adjoint matrix of $\bar{G}(\theta)$, whose elements are polynomials of elements of $\bar{G}(\theta)$.
It is trivial to see $\sup_\theta ||\bar{G}(\theta)|| < \infty$.
So $\sup_\theta ||\text{adj}(\bar{G}(\theta))|| < \infty$.
It follows immediately that $\sup_\theta ||\bar{G}(\theta)^{-1}|| < \infty$ as $\inf_\theta \det(\bar{G}(\theta)) > 0$.
\end{proof}

\subsection{Proof of Proposition~\ref{prop:tabular_mop}}
\begin{lemma} 
\label{lem:bertsekas}
(Corollary 6.1 in page 150 of \citet{bertsekas1989parallel})
If $Y$ is a square nonnegative matrix and $\rho(Y) < 1$, then there exists some vector $w \succ 0$ such that $||Y||^w_\infty < 1$.
Here $\succ$ is elementwise greater and $\rho(\cdot)$ is the spectral radius.
For a vector $y$, its $w$-weighted maximum norm is $||y||^w_\infty \doteq \max_i |\frac{y_i}{w_i}|$.
For a matrix $Y$, $||Y||^w_\infty \doteq \max_{y \neq 0} \frac{||Yy||^w_\infty}{||y||^w_\infty}$.
\end{lemma}
\textbf{Proposition~\ref{prop:tabular_mop}.}
\emph{$\mop$ is a contraction mapping w.r.t.\ some weighted maximum norm and $m_\pi$ is its fixed point.}
\begin{proof}
$\mop m_\pi = m_\pi$ follows directly from Eq~\eqref{eq:m_fixed_point}.
Given any two square matrices $A$ and $B$, 
the products $AB$ and $BA$ have the same eigenvalues (see Theorem 1.3.22 in \citet{horn2012matrix}).
Therefore $\rho(\gamma D^{-1}P_\pi^\top D) = \rho\big((\gamma P_\pi^\top D)D^{-1}\big) = \rho(\gamma P_\pi^\top) = \rho(\gamma P_\pi) < 1$.
Clearly $\gamma D^{-1}P_\pi^\top D$ is a nonnegative matrix.
Then Lemma~\ref{lem:bertsekas} implies that $\mop$ is a contraction mapping w.r.t. some weighted maximum norm.
\end{proof}

\subsection{Proof of Theorem~\ref{prop:convergence_GEM}}
\textbf{Theorem~\ref{prop:convergence_GEM}.}\emph{    (Convergence of GEM) Under Assumptions (\ref{assum:mdp}, \ref{assu:non_singular}) and Condition~\ref{con:actor_change_rate}, the iterate $\{d_t\}$ generated by \eqref{eq:gem-w} satisfies $\sup_t ||d_t|| < \infty$ and
    $\lim_{t \rightarrow \infty} ||\bar{G}(\theta_t)d_t - \bar{h}|| = 0 $ 
    almost surely. }
\begin{proof}
\label{sec:proof_of_thm1}
We first rephrase Theorem 3.1 in \citet{konda2002thesis}, which is a general theorem about the tracking ability of stochastic approximation algorithms under a slowly changing linear system. 
The original theorem adopts a stochastic process taking value in a Polish space.
Here we rephrase it for our finite MDP setting.
\begin{theorem}
\label{thm:konda}
\citep{konda2002thesis} Let $\{Y_t\}$ be a Markov chain with a finite state space $\mathcal{Y}$ and a transition kernel $\py \in \mathbb{R}^{|\mathcal{Y}| \times |\mathcal{Y}|}$,
consider an iterate $\{d_t\}$ in $\mathbb{R}^n$ evolving according to
\begin{align}
d_{t+1} = d_t + \alpha_t(h_{\theta_t}(Y_t) - G_{\theta_t}(Y_t)d_t),
\end{align}
where $\{\theta_t\}$ is another iterate in $\mathbb{R}^m$, $h_\theta: \mathcal{Y} \rightarrow \mathbb{R}^n$ and $G_\theta: \mathcal{Y} \rightarrow \mathbb{R}^{n \times n}$ are vector-valued and matrix-valued functions parameterized by $\theta$, $\alpha_t$ is a positive step size,
assume~\footnote{
\citet{konda2002thesis} considers a $\theta$-parameterized transition kernel $P_\mathcal{Y}^\theta$,
which changes every time step as $\theta_t$ evolves.
Here we consider a fixed transition kernel,
which is a special parameterization of $P_\mathcal{Y}^\theta$, 
i.e., we define $P_\mathcal{Y}^\theta \doteq P_\mathcal{Y}$ regardless of $\theta$.
The Assumption 3.1(1) in \citet{konda2002thesis} is then satisfied automatically.
The Assumptions \ref{assu:konda}(4,5) imply the Assumptions 3.1(7, 9) in \citet{konda2002thesis}.
}
\begin{assumption}
\label{assu:konda}
\quad \\
\begin{enumerate}
  \item (Stepsize) The sequence $\{\alpha_t\}$ is deterministic, non-increasing and satisfies the Robbins-Monro condition $$\sum_t \alpha_t = \infty, \quad \sum_t \alpha_t^2 < \infty$$
  \item (Parameter Changing Rate) The random sequence $\{\theta_t\}$ satisfies
  \begin{align}
   ||\theta_{t+1} - \theta_t|| \leq \beta_t H_t,
  \end{align}
  where $\{H_t\}$ is some nonnegative process with bounded moments and $\{\beta_t\}$ is a deterministic sequence such that $\sum_t (\frac{\beta_t}{\alpha_t})^d < \infty$ for some $d> 0$.
  \item (Poisson Equation) There exists $\hat{h}_\theta: \mathcal{Y} \rightarrow \mathbb{R}^n, \bar{h}(\theta) \in \mathbb{R}^n, \hat{G}_\theta: \mathcal{Y} \rightarrow \mathbb{R}^{n \times n}, \bar{G}(\theta) \in \mathbb{R}^{n \times n}$ such that
  \begin{align}
   \hat{h}_\theta(y) &= h_\theta(y) - \bar{h}(\theta) + \sum_{y^\prime} \py(y, y^\prime) \hat{h}_\theta(y^\prime), \\
   \hat{G}_\theta(y) &= G_\theta(y) - \bar{G}(\theta) + \sum_{y^\prime} \py(y, y^\prime) \hat{G}_\theta(y^\prime)
  \end{align}
  \item (Boundedness) There is a constant $C_0 < \infty$ such that $\forall \theta, y$,
  \begin{align}
  \max(|| \bar{h}(\theta) ||, || \bar{G}(\theta) ||, || \hat{h}_\theta(y) ||, || h_\theta(y) ||, || \hat{G}_\theta(y) ||, || G_\theta (y) ||) \leq C_0
  \end{align}
  \item (Lipschitz Continuity)
  There is a constant $C_0 < \infty$ such that $\forall \theta, \bar{\theta}, y$
  \begin{align}
  \max(||\bar{h}(\theta) - \bar{h}(\bar{\theta})||, ||\bar{G}(\theta) - \bar{G}(\bar{\theta})||) \leq C_0 || \theta - \bar{\theta} ||, \,
  ||f_\theta(y) - f_{\bar{\theta}}(y)|| \leq C_0 || \theta - \bar{\theta} ||
  \end{align}
  where $f_\theta$ represents any of $\hat{h}_\theta, h_\theta, \hat{G}_\theta, G_\theta$.
  \item (Uniformly Positive Definite) There exists a $\eta_0 > 0$ such that $\forall \theta, d,$
  \begin{align}
   d^\top \bar{G}(\theta) d \geq \eta_0 ||d||^2.
  \end{align}
\end{enumerate}
\end{assumption}
then~\footnote{The boundedness is from Lemma 3.9 in \citet{konda2002thesis}}
\begin{align}
\sup_t ||d_t|| < \infty, \quad \lim_t ||\bar{G}(\theta_t)d_t - \bar{h}(\theta_t)|| = 0 \quad a.s.
\end{align}
\end{theorem}
We now prove Theorem~\ref{prop:convergence_GEM} by verifying that GEM indeed satisfies Assumptions~\ref{assu:konda}(1-6).
Assumption~\ref{assu:konda}(1) is satisfied by the requirement of $\{\alpha_t\}$ in GEM. 
Assumption~\ref{assu:konda}(2) is satisfied by Condition~\ref{con:actor_change_rate}.

We now verify Assumption~\ref{assu:konda}(3).
Let 
\begin{align}
\mathcal{Y} \doteq \{(s, a, s') \mid s \in \mathcal{S}, a \in \mathcal{A}, s' \in \mathcal{S}, p(s'|s, a) > 0\},
\end{align}
$Y_t \doteq (S_t, A_t, S_{t+1}), y \doteq (s, a, s^\prime) \in \mathcal{Y}$ and $P_\mathcal{Y}$ be the transition kernel of $\{Y_t\}$, 
it is easy to verify that $P_\mathcal{Y}((s_1, a_1, s_1^\prime), (s_2, a_2, s_2^\prime)) = \mathbb{I}_{s_1^\prime = s_2} \mu(a_2 | s_2)p(s_2^\prime | s_2, a_2)$.
According to Assumption~\ref{assum:mdp}, the chain $\{Y_t\}$ is ergodic.
Let $\dy \in \mathbb{R}^{|\mathcal{Y}|}$ be its stationary distribution, we have
$\dy(y) = d_\mu(s)\mu(a|s)p(s^\prime|s, a)$.
For two fixed integers $i$ and $j$ in $[1, 2K_1]$, 
we consider an Markov Reward Process (MRP) with the state space $\mathcal{Y}$, 
the transition kernel $\py$ and 
the reward function $G_{\theta, ij}: \mathcal{Y} \rightarrow \R$, 
where $G_{\theta, ij}(y)$ is the $(i,j)$-indexed element in the matrix $G_\theta(y)$ (defined in Eq~\eqref{eq:GEM_G}).
Alternatively, we can view $G_{\theta, ij}$ as a vector in $\R^{|\mathcal{Y}|}$.
The average reward of this MRP is then $\bar{G}_{ij}(\theta)$, the $(i, j)$-indexed element in $\bar{G}(\theta)$ (defined in Eq~\eqref{eq:GEM_G_bar}).
We consider the differential value function (see \citet{sutton2018reinforcement}) $\hat{G}_{\theta, ij} \in \mathbb{R}^{|\mathcal{Y}|}$ of this MRP, where
\begin{align}
\hat{G}_{\theta, ij}(y) \doteq \mathbb{E}[\sum_{t=0}^\infty \big( G_{\theta, ij}(Y_t) - \bar{G}_{ij}(\theta) \big) | Y_0 = y].
\end{align}
These differential value functions define a matrix-valued function $\hat{G}_\theta: \mathcal{Y} \rightarrow \mathbb{R}^{2K_1 \times 2K_1}$.
Assumption~\ref{assu:konda}(3) is then satisfied according to the Bellman equation of differential value function (see \citet{sutton2018reinforcement}).
Moreover, according to the standard Markov chain theory (e.g., Section 8.2.1 in \citealt{puterman2014markov}),
we have 
\begin{align}
\hat{G}_{\theta, ij} = H_\mathcal{Y} G_{\theta, ij},
\end{align}
where $H_\mathcal{Y} \doteq (I - \py + \py^*)^{-1}(I - \py^*) \in \mathbb{R}^{|\mathcal{Y}| \times |\mathcal{Y}|}$ is the fundamental matrix which depends only on $\mu$ and $p$. 
Here each row of $\py^*$ is $\dy$.
Similarly, we define $\hat{h}: \mathcal{Y} \rightarrow \mathbb{R}^{2K_1}$ by defining its $i$-th component as
\begin{align}
\hat{h}_{i} \doteq H_\mathcal{Y} h_{i}.
\end{align}
Now $\hat{h}$, $h$ (defined in Eq~\eqref{eq:GEM_G}), and $\bar{h}$ (defined in Eq~\eqref{eq:GEM_G_bar}) satisfy Assumption~\ref{assu:konda}(3).
Note they are independent of $\theta$.

It is trivial to see that $\sup_\theta ||\bar{G}(\theta)|| < \infty$ and $||\bar{G}(\theta)||$ is Lipschitz continuous in $\theta$.
According to Assumption~\ref{assum:mdp}, $\rho_\theta$ is bounded.
It follows easily that $||\sup_\theta G_\theta(y)|| < \infty$. 
As $\mathcal{Y}$ is finite, $\sup_{\theta, y}||G_\theta(y)|| < \infty$.
Similarly, $||G_\theta(y)||$ is Lipschitz in $\theta$ and the Lipschitz constant is independent on $y$. 
Assumptions~\ref{assu:konda}(4, 5) are now satisfied.

We now verify Assumption~\ref{assu:konda}(6).
For any $d^\top = [\kappa^\top, w^\top]$, it is easy to verify that $$d^\top \bar{G}(\theta) d = \kappa^\top C \kappa + \eta w^\top w.$$
According to Assumption~\ref{assu:non_singular}, for any $\kappa \neq 0$, there exists a $\bar{\eta} > 0$ such that $\kappa^\top C \kappa \geq \bar{\eta}||\kappa||^2$, 
we therefore have 
\begin{align}
\label{eq:uniform_pd}
d^\top \bar{G}(\theta)d \geq \min(\bar{\eta}, \eta) ||d||^2, \quad \forall \theta, d.
\end{align}
Assumption~\ref{assu:konda}(6) is then satisfied.
Note here it is important to have $\eta > 0$.
If $\eta=0$ as in the original GTD2, $d^\top \bar{G}(\theta)d$ will not depend on $w$. 
Consequently, we can increase $||d||$ arbitrarily while keeping $d^\top \bar{G}(\theta)d$ unchanged by increasing $||w||$.
Assumption~\ref{assu:konda}(6) then cannot be satisfied.
\end{proof}

\subsection{Proof of Proposition~\ref{prop:linear_mop}}
\textbf{Proposition~\ref{prop:linear_mop}.}\emph{    $\Pi\mop(Xw_\theta^*(0)) = Xw_\theta^*(0)$. }
\begin{proof}
    Let $b \doteq X^\top D i$ and use similar techniques as \citet{hallak2017consistent},
    we have
    \begin{align}
        \Pi \mop (Xw_\theta^*(0)) &= X(X^\top D X)^{-1}X^\top D \Big(i + \gamma D^{-1}P_\pi^\top D Xw_\theta^*(0) \Big) \\
        &= X(X^\top D X)^{-1}b + \gamma X(X^\top D X)^{-1}X^\top P_\pi^\top D X A(\theta)^{-1}b \\
        &= X(X^\top D X)^{-1} \Big(A(\theta) + \gamma X^\top P_\pi^\top D X \Big) A(\theta)^{-1}b \\
        &= X(X^\top D X)^{-1} X^\top D X A(\theta)^{-1}b \quad \text{(Definition of $A(\theta)$)} \\
        &= Xw_\theta^*(0).
    \end{align}
\end{proof}

\subsection{Proof of Theorem~\ref{prop:convergence_GQ}}
\textbf{Theorem~\ref{prop:convergence_GQ}.} \emph{    (Convergence of GQ2) Under Assumptions (\ref{assum:mdp}, \ref{assu:non_singular}) and Condition~\ref{con:actor_change_rate}, the iterate $\{\tilde{d}_t\}$ generated by \eqref{eq:gq-w} satisfies $\sup_t || \tilde{d}_t || < \infty$ and
    $\lim_{t \rightarrow \infty} ||\tilde{G}(\theta_t)\tilde{d}_t - \tilde{h}|| = 0$ 
    almost surely.}
\begin{proof}
The proof is the same as the proof of Theorem~\ref{prop:convergence_GEM} up to a change of notations.
\end{proof}

\subsection{Proof of Theorem~\ref{thm:cof-pac}}
\textbf{Theorem~\ref{thm:cof-pac}.}\emph{    (Convergence of COF-PAC) Under Assumptions (\ref{assum:mdp}-\ref{assu:policy_params}), the iterate $\{\theta_t\}$ generated by COF-PAC (Algorithm~\ref{alg:cof-pac}) satisfies
    \begin{align}
        \lim \textstyle{\inf_t} \Big( ||\nabla J(\theta_t)|| - ||b(\theta_t)|| \Big) \leq 0,
    \end{align}
    almost surely, i.e., $\{\theta_t\}$ visits any neighborhood of the set
    $\{\theta : ||\nabla J(\theta)|| \leq ||b(\theta)|| \}$
    infinitely many times almost surely.}
\begin{proof}
This proof is inspired by \citet{konda2002thesis}.
With $\psi_t \doteq \rho_t \nabla \log \pi(A_t | S_t), w^*_t \doteq w^*_{\theta_t}(\eta), u^*_t \doteq u^*_{\theta_t}(\eta)$,
we rewrite the update $\Gamma_1(w_t) \Gamma_2(u_t) \Delta_t$ as
\begin{align}
\Gamma_1(w_t) \Gamma_2(u_t) \Delta_t &= \Gamma_1(w_t) (w_t^\top x_t) \Gamma_2(u_t) (u_t^\top \tilde{x}_t) \psi_t = e^{(1)}_t + e^{(2)}_t + g^*_t,
\end{align}
where 
\begin{align}
e^{(1)}_t &\doteq (\Gamma_1(w_t)w_t - \Gamma_1(w^*_t) w^*_t)^\top x_t \Gamma_2(u_t) u_t^\top \tilde{x}_t \psi_t, \\
e^{(2)}_t &\doteq \Gamma_1(w^*_t) {w^*_t}^\top x_t (\Gamma_2(u_t)u_t^\top - \Gamma_2(u^*_t) u^*_t)^\top \tilde{x}_t \psi_t, \\
g^*_t &\doteq \Gamma_1(w^*_t)(x_t^\top w^*_t) \Gamma_2(u^*_t) (\tilde{x}_t^\top u_t^*) \psi_t.
\end{align}
Then 
\begin{align}
    \theta_{t+1} &= \theta_t + \beta_t \Gamma_1(w_t) \Gamma_2(u_t) \Delta_t \\
    &= \theta_t + \beta_t e^{(1)}_t + \beta_t e^{(2)}_t + \beta_t g^*_t + \beta_t \Gamma_1(w^*_t) \Gamma_2(u^*_t) \Big(\nabla J(\theta_t) - \hat{g}(\theta_t) - b(\theta_t) \Big)
\end{align}
Using the second order Taylor expansion and Cauchy-Schwarz inequality, we have
\begin{align}
\label{eq:J_taylor}
    J(\theta_{t+1}) \geq &J(\theta_t) + \beta_t \Gamma_1(w^*_t) \Gamma_2(u^*_t) ||\nabla J(\theta_t)|| \Big( ||\nabla J(\theta_t)|| - ||b(\theta_t)|| \Big) \\
    &+ \beta_t \nabla J(\theta_t)^\top (g^*_t - \Gamma_1(w^*_t) \Gamma_2(u^*_t) \hat{g}(\theta_t)) \\
    &+ \beta_t \nabla J(\theta_t)^\top e_t^{(1)} \\
    &+ \beta_t \nabla J(\theta_t)^\top e_t^{(2)} \\
    &- \frac{1}{2} C_0 ||\beta_t \Gamma_1(w_t) \Gamma_2(u_t) \Delta_t||^2,
\end{align}
where $C_0$ reflects the bound of the Hessian.
We will prove in following subsections that all noise terms in Eq~\eqref{eq:J_taylor} are negligible. Namely,
\begin{lemma}
\label{lem:J_noise_1}
$\lim_{t \rightarrow \infty} e_t^{(i)} = 0 \quad a.s. \quad (i=1, 2)$
\end{lemma}
\begin{lemma}
\label{lem:J_noise_2}
$\sum_t ||\beta_t \Gamma_1(w_t) \Gamma_2(u_t) \Delta_t||^2$ converges a.s.
\end{lemma}
\begin{lemma}
\label{lem:J_noise_3}
$\sum_t \beta_t \nabla J(\theta_t)^\top (g^*_t - \Gamma_1(w^*_t) \Gamma_2(u^*_t) \hat{g}(\theta_t))$ converges a.s.
\end{lemma}

Same as the section ``Proof of Theorem 5.5'' in \citet{konda2002thesis}, we now consider a sequence $\{k_i\}$ such that
\begin{align}
k_0 = 0, \quad k_{i+1} = \min \{k \geq k_i\ | \sum_{l=k_i}^k \beta_l \geq T\} \, (i > 0)
\end{align}
for some constant $T > 0$.
Iterating Eq~\eqref{eq:J_taylor} yields
\begin{align}
\label{eq:last_J}
J(\theta_{k_{i+1}}) \geq J(\theta_{k_i}) + \delta_i + \sum_{t=k_i}^{k_{i+1}-1} \beta_t \Gamma_1(w^*_t) \Gamma_2(u^*_t) ||\nabla J(\theta_t)|| \Big( ||\nabla J(\theta_t)|| - ||b(\theta_t)|| \Big),
\end{align}
where
\begin{align}
\delta_i \doteq \sum_{t=k_i}^{k_{i+1}-1} \Big[ \beta_t \nabla J(\theta_t)^\top (g^*_t - \Gamma_1(w^*_t) \Gamma_2(u^*_t) \hat{g}(\theta_t)) + \beta_t \nabla J(\theta_t)^\top (e_t^{(1)} + e_t^{(2)}) - \frac{1}{2} C_0 ||\beta_t \Gamma_1(w_t) \Gamma_2(u_t) \Delta_t||^2 \Big].
\end{align}
Lemmas (\ref{lem:boundess_policy_params} - \ref{lem:J_noise_3}) and the selection of $\{k_i\}$ imply $\lim_{i \rightarrow \infty} \delta_i = 0$.
Theorems~\ref{prop:convergence_GEM} and~\ref{prop:convergence_GQ} imply that $w^*_t$ and $u^*_t$ are bounded. 
Consequently, $\Gamma_1(w^*_t)$ and $\Gamma_2(w^*_t)$ are bounded below by some positive constant.
If $\lim \inf_t \Big[ ||\nabla J(\theta_t)|| - ||b(\theta_t)|| \Big] \leq 0$ does not hold, 
there must exist $t_0$ and $\epsilon > 0$ such that $\forall t > t_0, \, ||\nabla J(\theta_t)|| - ||b(\theta_t)|| > \epsilon$.
So the summation in Eq~\eqref{eq:last_J} is bounded below.
Iterating Eq~\eqref{eq:last_J} then implies $J(\theta_t)$ is unbounded, which is impossible as $\gamma < 1$ and $r$ is bounded.
\end{proof}

\subsubsection{Proof of Lemma~\ref{lem:J_noise_1}}
\begin{proof}
It is easy to verify that $\Gamma_i$ is continuous.
According to Assumptions~(\ref{assu:non_singular}, \ref{assu:policy_params}), $x_t, \tilde{x}_t, \psi_t$ are bounded.
The reset follows immediately from Eq~\eqref{eq:gem_convergence} and Eq~\eqref{eq:gq_convergence}.
\end{proof}
\subsubsection{Proof of Lemma~\ref{lem:J_noise_2}}
\begin{proof}
It follows from the proof of Lemma~\ref{lem:J_noise_1} that $|| \Gamma_1(w_t)\Gamma_2(u_t)\Delta_t ||$ is bounded by some constant $C_0$.
Then $\sum_t ||\beta_t \Gamma_1(w_t)\Gamma_2(u_t)\Delta_t||^2 \leq C_0^2 \sum_t \beta_t^2 < \infty$.
\end{proof}
\subsubsection{Proof of Lemma~\ref{lem:J_noise_3}}
\begin{proof}
In this subsection we write $w_\theta^*(\eta)$ as $w_\theta^*$ for simplifying notation and define
\begin{align}
\label{eq:g_star_definition}
g^*_\theta(s, a) \doteq \Gamma_1(w_\theta^*) \Gamma_2(u_\theta^*) (x(s)^\top w_\theta^*) (\tilde{x}(s)^\top u_\theta^*) \psi_{\theta}(s, a).
\end{align}
So $g^*_t \doteq g^*_{\theta_t}(S_t, A_t)$.
Assumptions~(\ref{assu:non_singular},\ref{assu:policy_params}) imply that there is a constant $C_0 < \infty$ such that
\begin{align}
\forall (\theta, s, a), \, ||g^*_\theta(s, a)|| < C_0.
\end{align}
We first make a transformation of the original noise using Poisson equation as in Section~\ref{sec:proof_of_thm1}.
Similarly, let $\mathcal{Y} \doteq \mathcal{S} \times \mathcal{A}, Y_t \doteq (S_t, A_t), y \doteq (s, a)$ and $\py$ be the transition kernel of $\{Y_t\}$, 
i.e., $\py \big( (s, a), (\p{s}, \p{a}) \big) \doteq p(\p{s}|s, a)\mu(\p{a}|\p{s})$.
For every integer $i$ in $[1, K]$, we consider the MRP with the reward function $g^*_{\theta,i}: \mathcal{Y} \rightarrow \R$, 
where $g^*_{\theta, i}(y)$ is the $i$-the element of $g^*_\theta(y)$.
Again, we view $g^*_{\theta, i}$ as a vector in $\R^{|\mathcal{Y}|}$.
The average reward is therefore $\bar{g}_i(\theta)$, which is the $i$-th element of $\bar{g}(\theta) \doteq \Gamma_1(w^*_\theta) \Gamma_2(u^*_\theta) \hat{g}(\theta) \in \R^K$.
With $H_\mathcal{Y}$ to denoting the fundamental matrix of this MRP,
the differential value function of this MRP is then
\begin{align}
\label{eq:v_hat_definition}
\hat{v}_{\theta, i} \doteq H_\mathcal{Y} g^*_{\theta, i} \in \R^{|\mathcal{Y}|}.
\end{align}
These differential value functions define a vector-valued function $\hat{v}_\theta: \mathcal{Y} \rightarrow \R^K$, 
which satisfies $\sup_{\theta, s, a} ||\hat{v}_\theta(s, a)|| < \infty$
as it is just linear transformation of $g_\theta^*$.
According to the Bellman equation of differential value function, we have
\begin{align}
\label{eq:v_hat_bellman}
\hat{v}_\theta(y) = g^*_\theta(y) - \bar{g}(\theta) + \sum_{\p{y}}\py(y, \p{y})\hat{v}_\theta(\p{y}).
\end{align}

Now we are ready to decompose the noise $\nabla J(\theta_t)^\top (g^*_t - \Gamma_1(w^*_t) \Gamma_2(u^*_t) \hat{g}(\theta_t))$ as
\begin{align}
    &\nabla J(\theta_t)^\top (g^*_t - \Gamma_1(w^*_t) \Gamma_2(u^*_t) \hat{g}(\theta_t)) \\
    =& \nabla J(\theta_t)^\top (g^*_{\theta_t}(S_t, A_t) - \bar{g}(\theta_t)) \quad \text{(Definition of $g^*_{\theta_t}$ and $\bar{g}(\theta_t)$)} \\
    =& \nabla J(\theta_t)^\top \Big(\hat{v}_{\theta_t}(S_t, A_t) - \sum_{s^\prime, a^\prime} p(s^\prime|S_t, A_t)\mu(a^\prime|s^\prime) \hat{v}_{\theta_t}(s^\prime, a^\prime) \Big) \quad \text{(Eq~\eqref{eq:v_hat_bellman})}\\
    =& \sum_{i=1}^4 \epsilon^{(i)}_t,
\end{align}
where
\begin{align}
    \epsilon^{(1)}_t &\doteq \nabla J(\theta_t)^\top \Big(\hat{v}_{\theta_t}(S_{t+1}, A_{t+1}) - \sum_{s^\prime, a^\prime} p(s^\prime|S_t, A_t)\mu(a^\prime|s^\prime) \hat{v}_{\theta_t}(s^\prime, a^\prime) \Big), \\
    \epsilon^{(2)}_t &\doteq \frac{\beta_{t-1} \nabla J(\theta_{t-1})^\top \hat{v}_{\theta_{t-1}}(S_t, A_t) - \beta_t \nabla J(\theta_t)^\top \hat{v}_{\theta_t} (S_{t+1}, A_{t+1})}{\beta_t}, \\
    \epsilon^{(3)}_t &\doteq \frac{\beta_t - \beta_{t-1}}{\beta_t} \nabla J(\theta_{t-1})^\top \hat{v}_{\theta_{t-1}}(S_t, A_t), \\
    \epsilon^{(4)}_t &\doteq \nabla J(\theta_t)^\top \hat{v}_{\theta_t}(S_t, A_t) - \nabla J(\theta_{t-1})^\top \hat{v}_{\theta_{t-1}}(S_t, A_t). \\
\end{align}
We now show $\sum_t \beta_t \epsilon_t^{(i)}$ converges almost surely for $i=1, 2, 3, 4$.

(1)
We first state a Martingale Convergence Theorem (see Proposition 4.3 in \citealt{bertsekas1996neuro}).
\begin{lemma}
\label{lem:martingale_convergence}
Assuming $\{M_l\}_{l=1,\dots}$ is a Martingale sequence and there exists a constant $C_0 < \infty$ such that
$\forall l, \mathbb{E}[|M_l|^2] < C_0$, then $\{M_l\}$ converges almost surely.
\end{lemma}
Let $\mathcal{F}_l \doteq \sigma(S_0, A_0, \theta_0, \dots, S_l, A_l, \theta_l, S_{l+1}, A_{l+1})$ be the $\sigma$-algebra
and $M_l \doteq \sum_{t=0}^l \beta_t \epsilon_t^{(1)}$. 
It is easy to see that $M_l$ is adapted to $\mathcal{F}_l$. 
Due to Lemma~\ref{lem:boundess_policy_params} and boundedness of $\hat{v}$, $|\epsilon^{(1)}_t| < C_1$ for some constant $C_1$, 
implying $\mathbb{E}[|M_l|] < \infty$ holds for any fixed $l$.
Moreover,
\begin{align}
    \mathbb{E}[M_{l+1} | \mathcal{F}_l] &= M_l + \mathbb{E}_{\theta_{l+1}, S_{l+2}, A_{l+2}}[\beta_{l+1} \epsilon_{l+1}^{(1)}| \mathcal{F}_l] \\
    &= M_l + \beta_{l+1} \mathbb{E}_{\theta_{l+1}} \Big[ \mathbb{E}_{S_{l+2}, A_{l+2}}[ \epsilon^{(1)}_{l+1} |\theta_{l+1}, \mathcal{F}_l] \Big] \\
    &= M_l + \beta_{l+1} \mathbb{E}_{\theta_{l+1}} [ 0 ] = M_l
\end{align}
$M_l$ is therefore a Martingale.
We now verify that $M_l$ has bounded second moments, then $\{M_l\}$ converges according to Lemma~\ref{lem:martingale_convergence}. 
For any $t_1 < t_2$, we have
\begin{align}
    \mathbb{E}[\epsilon_{t_1}^{(1)}\epsilon_{t_2}^{(1)}] = \mathbb{E} \Big[ \mathbb{E}[\epsilon_{t_1}^{(1)}\epsilon_{t_2}^{(1)} | \mathcal{F}_{t_2 - 1} ] \Big] = \mathbb{E} \Big[\epsilon_{t_1}^{(1)} \mathbb{E}[\epsilon_{t_2}^{(1)} | \mathcal{F}_{t_2 - 1} ] \Big]
    = \mathbb{E} \Big[\epsilon_{t_1}^{(1)} 0 \Big] = 0.
\end{align}
Consequently,
\begin{align}
    \forall l, \quad \mathbb{E}[|M_l|^2] &= \mathbb{E}[\sum_{t=0}^l \beta_t^2 \big( \epsilon^{(1)}_t \big)^2] \leq C_1^2 \sum_{t=0}^\infty \beta^2_t.
\end{align}
Therefore, $\{M_l\}$ and $\sum_t \beta_t \epsilon^{(1)}_t$ converges a.s..

(2) $\sum_{t=1}^l \beta_t \epsilon^{(2)}_t = \beta_0 \nabla J(\theta_0)^\top \hat{v}_{\theta_0}(S_1, A_1) - \beta_l \nabla J(\theta_l)^\top \hat{v}_{\theta_l}(S_{l+1}, A_{l+1})$. 
The rest follows from the boundedness of $\nabla J(\theta)$ and $\hat{v}_\theta(s, a)$ and $\lim_{l\rightarrow \infty} \beta_l = 0$.

(3) 
\begin{align}
\sum_{t=1}^l |\beta_t \epsilon^{(3)}_t| &\leq \sum_{t=1}^l |\beta_t - \beta_{t-1}| \, |\nabla J(\theta_{t-1})^\top \hat{v}_{\theta_{t-1}}(S_t, A_t)| \\
&\leq C_1 \sum_{t=1}^l (\beta_{t-1} - \beta_t) \leq C_1 (\beta_0 - \beta_l) < C_1 \beta_0 \quad a.s.
\end{align}
It follows easily that $\sum_t \beta_t \epsilon_t^{(3)}$ converges absolutely, thus converges.

(4) Lemma~\ref{lem:boundess_policy_params} implies $\nabla J(\theta)$ is bounded and Lipschitz continuous in $\theta$, 
if we are able to show $\forall (s, a, t)$, there exists a constant $C_0$ such that
\begin{align}
\label{eq:lipschitz_v_hat}
||\hat{v}_{\theta_t}(S_t, A_t) - \hat{v}_{\theta_{t-1}}(S_t, A_t)|| \leq C_0 ||\theta_t - \theta_{t-1}||,
\end{align}
we will have
\begin{align}
|\epsilon^{(4)}_t| &\leq C_1 ||\theta_t - \theta_{t-1}|| = C_1 ||\beta_t \Gamma_1(w_t)\Gamma_2(u_t)\Delta_t|| \leq \beta_t C_2.
\end{align}
Consequently,
\begin{align}
\sum_{t=1}^l |\beta_t \epsilon^{(4)}_t| &\leq C_2 \sum_{t=1}^l \beta_t^2 < C_2 \sum_{t=1}^\infty \beta_t^2 \quad a.s.
\end{align}
Thus $\sum_t \beta_t \epsilon_t^{(4)}$ converges.
We now proceed to show Eq~\eqref{eq:lipschitz_v_hat} does hold. 
According to Eq~\eqref{eq:v_hat_definition}, it suffices to show $\forall (s, a, \theta, \bar{\theta})$, there exists a constant $C_0$ such that
\begin{align}
|| g^*_{\theta}(s, a) - g^*_{\bar{\theta}}(s, a) || \leq C_0 || \theta - \bar{\theta}||.
\end{align}
According to Assumption~\ref{assu:policy_params}, $\psi_\theta(s, a)$ is bounded and Lipschitz continuous in $\theta$. 
It is easy to verify the function $||w\Gamma_i(w)||$ is bounded and Lipschitz continuous in $w$ in a compact set.
According to the definition of $g^*_\theta(s, a)$ in Eq~\eqref{eq:g_star_definition} and the boundedness of
$w^*_\theta$ and $u^*_\theta$,
it then suffices to show $w_\theta^*$ and $u_\theta^*$ are Lipschitz continuous in $\theta$. 
Recall by definition $w^*_{\theta}$ is the second half of $\bar{G}(\theta)^{-1} \bar{h}$,
it then suffices to show $\bar{G}(\theta)^{-1}$ is Lipschitz continuous in $\theta$. 
Using the fact 
\begin{align}
    \label{eq:the_fact}
    ||B_1^{-1} - B_2^{-1}|| = ||B_1^{-1}(B_1 - B_2) B_2^{-1}|| \leq ||B_1^{-1}|| \, ||B_1 - B_2|| \, ||B_2^{-1}||, 
\end{align}
we have
\begin{align}
||\bar{G}(\theta)^{-1} - \bar{G}(\bar{\theta})^{-1} || \leq ||\bar{G}(\theta)^{-1} || \, || \bar{G}(\theta) - \bar{G}(\bar{\theta})|| \, ||\bar{G}(\bar{\theta})^{-1} ||.
\end{align}
The rest follows from the fact that $\bar{G}(\theta)$ is Lipschitz continuous in $\theta$ and Lemma~\ref{lem:G_inverse}.
Similarly, we can establish the Lipschitz continuity of $u_\theta^*$, which completes the proof.
\end{proof}

\subsection{Proof of Proposition~\ref{prop:b_theta_bound}}
\textbf{Proposition~\ref{prop:b_theta_bound}.}
\emph{Under Assumptions~(\ref{assum:mdp}-\ref{assum:bound_of_bias}), 
let $d_\theta$ be the stationary distribution under $\pi_\theta$ and define $\tilde{d}_\theta(s, a) \doteq d_\theta(s) \pi_\theta(a|s), D_\theta\doteq diag(d_\theta), \tilde{D}_\theta \doteq diag(\tilde{d}_\theta)$,
we have
\begin{align}
||b(\theta)||_D \leq C_0\eta
&+ C_1 \textstyle{\frac{1 + \gamma \kappa(D^{-\frac{1}{2}} D_\theta^{\frac{1}{2}})}{1-\gamma}}
||m_{\pi_\theta} - \Pi m_{\pi_\theta}||_D \\
&+ C_2\textstyle{\frac{1 + \gamma \kappa(\tilde{D}^{-\frac{1}{2}} \tilde{D}_\theta^{\frac{1}{2}})}{1-\gamma}} ||q_{\pi_\theta} - \tilde{\Pi} q_{\pi_\theta}||_{\tilde{D}},
\end{align}
where $\kappa(\cdot)$ is the condition number of a matrix w.r.t.\ $\ell_2$ norm and $C_0, C_1, C_2$ are some positive constants.}
\begin{proof}
We first decompose $b(\theta)$ as $b(\theta) = b_1(\theta) + b_2(\theta)$, where 
\begin{align}
b_1(\theta) &\doteq \sum_s d_\mu(s) \big(m_{\pi_\theta}(s) - x(s)^\top w^*_\theta(\eta) \big) \sum_a \mu(a|s) \psi_\theta(s, a) \big( \tilde{x}(s, a)^\top u^*_\theta(\eta) \big) \\
b_2(\theta) &\doteq \sum_s d_\mu(s) m_{\pi_\theta}(s) \sum_a \mu(a|s) \psi_\theta(s, a) \big(q_{\pi_\theta}(s, a) -  \tilde{x}(s, a)^\top u^*_\theta(\eta) \big)
\end{align}
The boundedness of $m_\pi(s)$ and $\psi_\theta(s, a)$ implies
\begin{align}
||b_2(\theta)||_D \leq C_2 ||q_{\pi_\theta} - \tilde{X}u^*_\theta(\eta)||_D \leq C_2 ||q_{\pi_\theta} - \tilde{X}u^*_\theta(0)||_D + C_2 || \tilde{X}u^*_\theta(\eta) - \tilde{X}u^*_\theta(0) ||_D
\end{align}
for some constant $C_2$. Theorem 2 in \citet{kolter2011fixed} states 
\begin{align}
||q_{\pi_\theta} - \tilde{X}u^*_\theta(0)||_D \leq \frac{1 + \gamma \kappa(\tilde{D}^{-\frac{1}{2}} \tilde{D}_\theta^{\frac{1}{2}})}{1-\gamma}||q_{\pi_\theta} - \tilde{\Pi} q_{\pi_\theta}||_{\tilde{D}}.
\end{align}
Moreover, we have
\begin{align}
&|| \tilde{X}u^*_\theta(\eta) - \tilde{X}u^*_\theta(0) ||_D \leq ||\tilde{X}||_D ||u^*_\theta(\eta) - u^*_\theta(0) ||_D \\
\leq& ||\tilde{X}||_D \, ||(\tilde{A}(\theta)^\top \tilde{C}^{-1} \tilde{A}(\theta) + \eta I)^{-1} ||_D \, \eta ||I||_D \, ||(\tilde{A}(\theta)^\top \tilde{C}^{-1} \tilde{A}(\theta))^{-1} ||_D \, ||\tilde{A}(\theta)^\top \tilde{C}^{-1} \tilde{X}^\top \tilde{D}\tilde{r} ||_D,
\end{align}
where the last inequality results from Eq~\eqref{eq:the_fact}.
The boundedness of $||(\tilde{A}(\theta)^\top \tilde{C}^{-1} \tilde{A}(\theta) + \eta I)^{-1} ||_D$ can be established with the same routine as the proof of Lemma~\ref{lem:G_inverse}.
The boundedness of $||(\tilde{A}(\theta)^\top \tilde{C}^{-1} \tilde{A}(\theta))^{-1} ||_D$ can be established with this routine and Assumption~\ref{assum:bound_of_bias}(b), 
which completes the boundedness of $b_2(\theta)$.

The same routine can be used to bound $b_1(\theta)$.
Particularly, we need to show
\begin{align}
||m_{\pi_\theta} - Xw^*_\theta(0) ||_D \leq \frac{1 + \gamma \kappa(D^{-\frac{1}{2}} D_\theta^{\frac{1}{2}})}{1-\gamma}||m_{\pi_\theta} - \Pi m_{\pi_\theta}||_D.
\end{align}
We omit the proof for this inequality as it is mainly verbatim repetition of the proof of Theorem 2 in \citet{kolter2011fixed} except two main differences:
(1) The positive semidefiniteness of $F_\theta$ in Assumption~\ref{assum:bound_of_bias} ensures $||\Pi D^{-1}P^\top_\pi D Xw||_D \leq ||Xw||_D \, \forall w$, analogously to Eq~(10) in \citet{kolter2011fixed}.
We will then replace the occurrence of $P_\pi$ in \citet{kolter2011fixed} with $D^{-1}P_\pi^\top D$.
(2) When we reach the place to compute $||D^{-1}P_\pi^\top D||_D$, analogously to Eq~(17) in \citet{kolter2011fixed},
Lemma~\ref{lem:p_pi_norm} implies we need only $||P_\pi||_D$, which has been computed by \citet{kolter2011fixed}.
\end{proof}

\section{Experiment Details}
We implemented the variant Baird's counterexample \citep{sutton2018reinforcement} by ourselves.
\texttt{Reacher-v2} is from Open AI gym \citep{brockman2016openai}~\footnote{\url{https://gym.openai.com/}}.
We conducted our experiments in a server with an Intel\textsuperscript{\textregistered} Xeon\textsuperscript{\textregistered} Gold 6152 CPU.
Our implementation is based on PyTorch.

\subsection{Original Features of Baird's Counterexample}
According to \cite{sutton2018reinforcement}, we have
\begin{align}
x(s_1) \doteq [2, 0, 0, 0, 0, 0, 0, 1]^\top \\
x(s_2) \doteq [0, 2, 0, 0, 0, 0, 0, 1]^\top \\
x(s_3) \doteq [0, 0, 2, 0, 0, 0, 0, 1]^\top \\
x(s_4) \doteq [0, 0, 0, 2, 0, 0, 0, 1]^\top \\
x(s_5) \doteq [0, 0, 0, 0, 2, 0, 0, 1]^\top \\
x(s_6) \doteq [0, 0, 0, 0, 0, 2, 0, 1]^\top \\
x(s_7) \doteq [0, 0, 0, 0, 0, 0, 1, 2]^\top \\
\end{align}

\subsection{Mujoco Experiments}
To evaluate $J(\pi)$, we first sample a state from $d_\mu$ and then follow $\pi$ until episode termination, which we call an excursion.
We use the averaged return of 10 excursions as an estimate of $J(\pi)$.

Our implementation is based on the open-sourced implementation from \citet{zhang2019generalized}.
We, therefore, refer the reader to \citet{zhang2019generalized} for full details.
In particular, two-hidden-layer networks are used to parameterize $\pi, v_\pi, m_\pi$,
and the target policy $\pi$ is a Gaussian policy.
We do not clip the importance sampling ratios.
We use a semi-gradient rule to learn $m_\pi$.
Assuming $m_\pi$ is parameterized by $\theta_m$,
we then update $\theta_m$ to minimize
\begin{align}
    \Big( i(S_{t+1}) + \gamma \rho(S_t, A_t) m_\pi(S_t; \bar{\theta}_m) - m_\pi(S_{t+1}; \theta_m) \Big)^2,
\end{align}
where $\bar{\theta}_m$ indicates the parameters of the target network.
Our COF-PAC implementation inherits the hyperparameters from the ACE implementation of \citet{zhang2019generalized} without further tuning.
Our implementation of TD3 with uniformly random behavior policy is taken from \citet{zhang2019generalized} directly.

\end{document}